\documentclass[11pt]{article}

\usepackage{acl}

\usepackage{times}
\usepackage{latexsym}

\usepackage[T1]{fontenc}

\usepackage[utf8]{inputenc}

\usepackage{microtype}

\usepackage{inconsolata}

\usepackage{graphicx}

\usepackage{multirow}
\usepackage{booktabs}
\usepackage{bm}
\usepackage{amsmath}
\usepackage{amssymb}
\usepackage{xcolor}
\usepackage{colortbl}
\usepackage{subfigure}

\title{UserLM-R1: Modeling Human Reasoning in User Language Models with Multi-Reward Reinforcement Learning}

\author{Feng Zhang$^{1,2}$\Thanks{Work done during internship at Meituan.} \quad Shijia Li$^{1,3 *}$ \quad  Chunmao Zhang$^{1,4 *}$ \quad  \\  {\bf  Zhanyu Ma$^{1}$  \quad Jun Xu$^{1}$\footnotemark[2] \quad  Jiuchong Gao$^{1}$\footnotemark[2]}\\ {\bf  Jinghua Hao$^{1}$ \quad Renqing He$^{1}$ \quad \quad Jingwen Xu$^{1}$ \quad  Han Liu$^{5}$\Thanks{Corresponding author.}} \\
$^1$Meituan, $^2$Peking University,
$^3$Beijing University of Posts and Telecommunications, \\
$^4$University of Chinese Academy of Sciences, $^5$Dalian University of Technology \\
}

\begin{document}
\maketitle
\begin{abstract}
User simulators serve as the critical interactive environment for agent post-training, and an ideal user simulator generalizes across domains and proactively engages in negotiation by challenging or bargaining. However, current methods exhibit two issues.  They rely on static and context-unaware profiles, necessitating extensive manual redesign for new scenarios, thus limiting generalizability. Moreover, they neglect human strategic thinking, leading to vulnerability to agent manipulation. To address these issues, we propose UserLM-R1, a novel user language model with reasoning capability. Specifically, we first construct comprehensive user profiles with both static roles and dynamic scenario-specific goals for adaptation to diverse scenarios. Then, we propose a goal-driven decision-making policy to generate high-quality rationales before producing responses, and further refine the reasoning and improve strategic capabilities with supervised fine-tuning and multi-reward reinforcement learning. Extensive experimental results demonstrate that UserLM-R1 outperforms competitive baselines, particularly on the more challenging adversarial set.
\end{abstract}

\section{Introduction}

\begin{figure}[t]
		\centering
		\includegraphics[width=7cm]{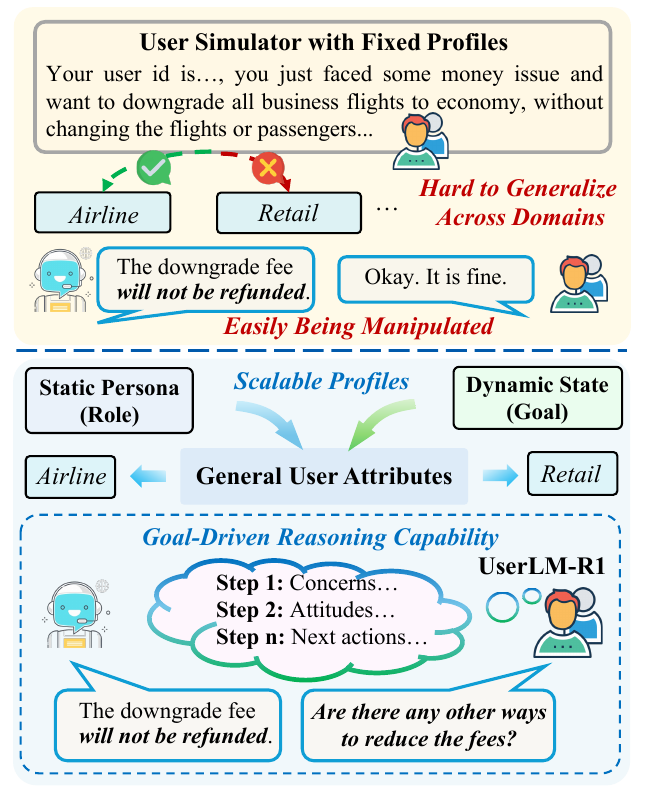}
		\caption{Comparison of previous methods and ours. }
		\label{paradigm}
\end{figure}

Large Language Models (LLMs) have revolutionized interaction tasks due to their semantic understanding and instruction-following capabilities \cite{DBLP:journals/corr/abs-2303-18223, DBLP:journals/corr/abs-2307-12966, DBLP:journals/fcsc/WangMFZYZCTCLZWW24}. As post-training becomes critical for optimizing LLM performance, scalable environment feedback is indispensable \cite{DBLP:journals/corr/abs-2502-21321}. To overcome the bottleneck of acquiring real human feedback, user simulators are proposed. User simulators aim to mimic diverse human users with different personas and behavior patterns in a scalable and reproducible manner \cite{DBLP:journals/tmlr/Chen00YZSXLYZCL24}. This emerging topic has played important roles in generating rollout data for task agent training \cite{DBLP:conf/acl/Liu0YWM0WW23, DBLP:journals/corr/abs-2504-03601, DBLP:journals/corr/abs-2509-19736}, evaluating agents with the controlled and reproducible manner \cite{DBLP:journals/corr/abs-2406-12045, DBLP:journals/corr/abs-2309-13233, DBLP:journals/tois/SunGZRCRR24, DBLP:conf/naacl/LuHZANBMMLYWP25}, and analyzing human behavior \cite{argyle2023out, DBLP:conf/uist/ParkOCMLB23}, which allows for large-scale interaction research. The fidelity of user simulation regarding human behavior and decisions is critical, as it directly impacts the trained policies and evaluation of task agents \cite{DBLP:conf/emnlp/ShiQWY19}. Good user simulators are able to generalize across diverse domains and proactively engage in negotiation and assert their interests, such as bargaining.

Recent studies mainly focus on role-driven and goal-driven user simulators. One aims to provide emotional value by maintaining a consistent and credible persona in interactions \cite{DBLP:conf/acl/WangL000025, DBLP:conf/acl/YuYWZQ25}, and the other regards solving practical problems as its core positioning and focuses on achieving goals \cite{DBLP:journals/corr/abs-2510-06552, DBLP:journals/corr/abs-2507-20152}. However, as illustrated in Figure \ref{paradigm}, existing studies suffer from two issues. On the one hand, predefined static role profiles are tailored for the specific task environment, requiring manual design for different scenarios, which greatly limits the generalization of the user simulator. On the other hand, they neglect the strategic thought process and the nuance of human decision-making, resulting in simulators that are prone to being manipulated. As shown in Figure \ref{paradigm}, the simulator lacks resistance, easily capitulating to the conditions of the agent when faced with a negative response.

To alleviate these problems, we propose UserLM-R1, a novel reasoning user language model with human-like strategic reasoning paths to keep responses consistent with their profiles and adhering to goal compliance. Specifically, to enhance the generalization, we first propose scalable user profiles that encompass both static personas and dynamic goals. The static persona serves as the stable foundation of the user, including background information, personality traits, expression style, and life scenarios. Conversely, dynamic goals represent the scenario-specific agenda and behaviors the user exhibits, which are generated dynamically by the user simulator during the conversation. Second, to improve the proactive strategy capabilities, we design a goal-driven reasoning strategy that mimics how humans make decisions. Based on this, we construct explicit reasoning traces and then conduct supervised fine-tuning on the generated conversation samples to foster intrinsic capabilities to think and make decisions before generating responses. Furthermore, we design composite rewards for reinforcement learning to allow the model to explore more diverse and logical reasoning trajectories.

In experiments, we evaluate user simulators from both session-level and turn-level perspectives, where a more challenging adversarial test set with 11 types of traps is introduced. Our evaluation metrics capture not only role consistency and goal progress but also the proactive strategic capacity. Experimental results demonstrate that UserLM-R1 significantly outperforms competitive baselines. Our main contributions are as follows:

\begin{itemize}
    \item We present \textbf{UserLM-R1}, a novel user simulator with reasoning capability. By mimicking the typical human thinking process, the reasoning covers the decision trajectories and state changes before generating responses.

    \item We propose a framework that creates generalizable user profiles with both static and dynamic attributes, further enhanced by goal-driven reasoning fine-tuning and multi-reward reinforcement learning, for better generalizability and stability in simulation.

    \item Comprehensive evaluation results show that UserLM-R1 outperforms baselines in both session and turn datasets and enables effective task agent training.
    
\end{itemize}

\begin{figure*}[t]
		\centering
		\includegraphics[width=1\linewidth]{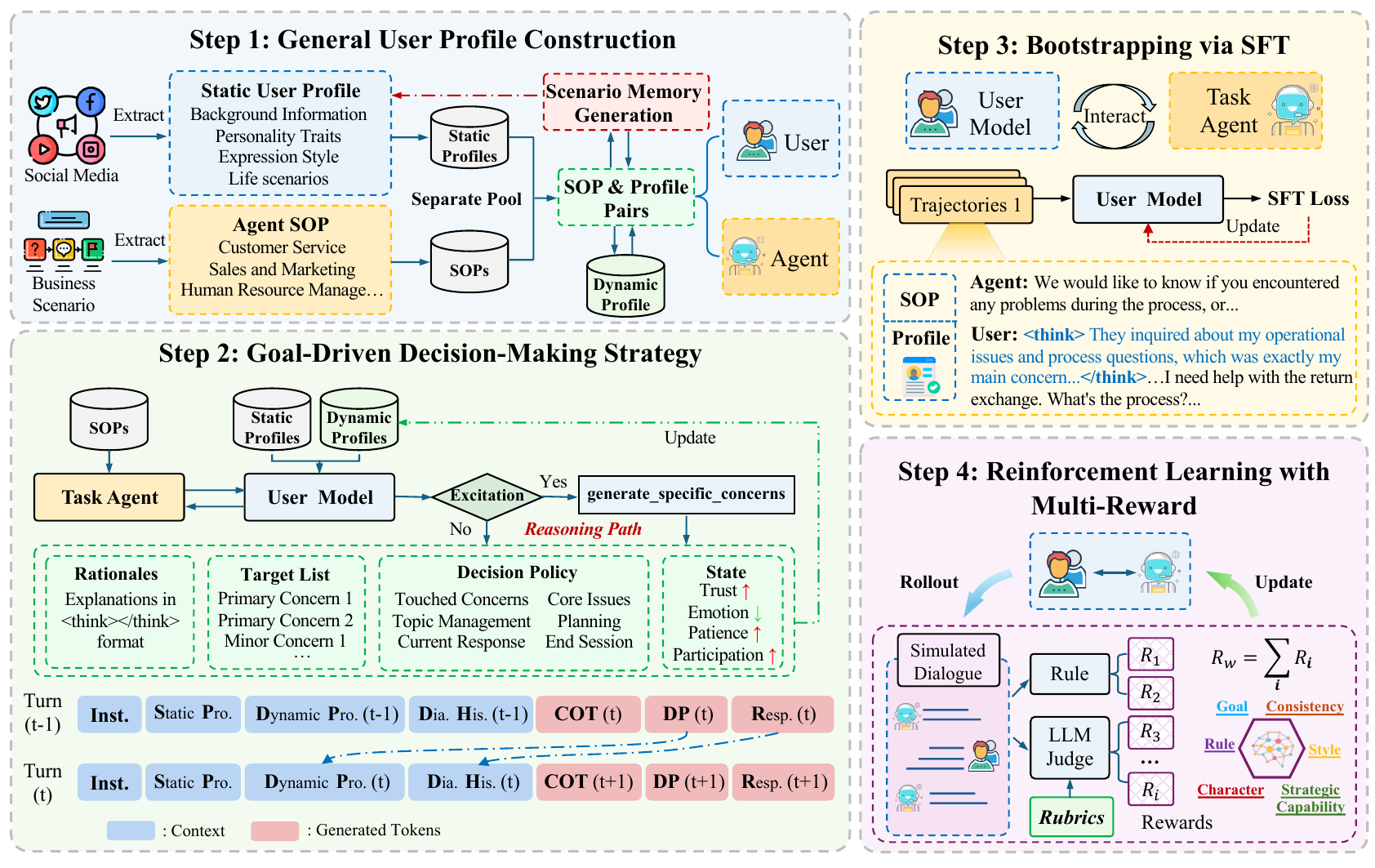}
		\caption{The overview of our proposed UserLM-R1. First, general user profiles are constructed. Second, the goal-driven decision-making strategy enables the user model to have reasoning capability. Finally, supervised fine-tuning and multi-reward reinforcement learning are used to further improve the proactive strategic ability.}
		\label{fw}
\end{figure*}

\section{Related Work}

\subsection{User Simulation}
User simulation aims to create a model that mimics the actions of human users with complex and variable individual preferences, which has attracted increasing attention \cite{DBLP:journals/corr/abs-2501-04410}. Role-driven simulation \cite{DBLP:conf/emnlp/Zhou0WWSYHKBPYX24, DBLP:journals/corr/abs-2401-09432, DBLP:conf/acl/WangPQLZWGGN00024} aims to conduct conversation with a predefined role, like style and character views, and is commonly applied in entertainment scenarios, like chit-chat. Beyond Dialogue \cite{DBLP:conf/acl/YuYWZQ25} leverages the crafted prompts to create a fine-grained alignment dataset to alleviate the biases and even conflicts between dialogue and profiles. Goal-driven simulation \cite{DBLP:journals/corr/abs-2402-13374} aims to achieve specific objectives in tasks with more underlying information needed, where the additional objective task information is appended. DAUS \cite{DBLP:journals/corr/abs-2402-13374} is fine-tuned on real examples of task-oriented dialogues, improving user goal fulfillment. UGST \cite{DBLP:journals/corr/abs-2507-20152} focuses on tracking the user goal state via a three-stage method. In this paper, we propose UserLM-R1, the first user simulator with the reasoning process covering decision trajectories and state changes.

\subsection{Reinforcement Learning for Reasoning}

Reinforcement learning has emerged as a promising paradigm to advance the reasoning capabilities of large language models for solving complex tasks, especially reinforcement learning with verifiable rewards (RLVR) \cite{DBLP:journals/corr/abs-2412-16720, DBLP:journals/corr/abs-2501-12948, DBLP:journals/corr/abs-2411-15124}. Most RLVR methods rely on rule-based verifiers to obtain rewards in the mathematics and code domains \cite{DBLP:journals/corr/abs-2502-01456} via GRPO \cite{DBLP:journals/corr/abs-2402-03300} or other optimization policy. Recent work \cite{DBLP:journals/corr/abs-2509-20357} states that language models that think, chat better, like in writing outline essays or making plans where humans reason routinely. For user simulation, it is natural to first generate a reasoning process before making decisions or providing responses, motivating us to develop UserLM-R1.

\subsection{User Simulator Evaluation}
The evaluation for user simulators can be divided into metric-based evaluation \cite{DBLP:conf/www/ChenDZFZT0ZD25}, model-based evaluation\cite{DBLP:conf/acl/YuYWZQ25}, and human evaluation \cite{DBLP:journals/corr/abs-2511-00222}. Metric-based methods primarily assess the diversity and authenticity using the designed metrics, like using n-grams and BLEU to evaluate the diversity. Model-based methods usually leverage LLMs as judges to evaluate the generated utterances and dialogues, for example. Human evaluation methods are more accurate but are costly and difficult to replicate. However, these previous studies focus on user-initiated scenarios and neglect the strategic capabilities of user simulators. In this paper, we construct the evaluation samples and metrics for user-passive scenarios. We introduce a more challenging adversarial dataset and propose evaluation metrics designed to assess resistance to manipulation and game-theoretic abilities.

\section{Task Formulation}

Formally, we define a dialogue with $n$ turns $\bm{d}_i = \{(\bm{a}_{i1}, \bm{u}_{i1}), (\bm{a}_{i2}, \bm{u}_{i2}), ..., (\bm{a}_{in}, \bm{u}_{in})\}$, where $\bm{a}_{i1}$ and $\bm{u}_{i1}$ denote the task agent and user utterance at the first turn, respectively. In this paper, we focus on the scenarios where the task agent initiates the dialogue first. The task agent is provided with an initial system message $C_i$, which provides general task instructions, including the task scenarios and the purpose of the conversation. The user simulator is provided with profiles $\bm{p} = (\bm{p}^s, \bm{p}^d)$, including the static user profile $\bm{p}^s$ and the dynamic user profile $\bm{p}^d$. The static user profile $\bm{p}^s$ forms the foundation of the user and represents a relatively stable and unchanging background setting. The dynamic user profile $\bm{p}^d$, which evolves throughout the dialogue, captures the specific goals and behaviors a user exhibits in a particular scenario. The combination of these two components enables the model to embody a real person with a past, personality, and current intentions, thereby enhancing authenticity and credibility. Additionally, this setup simulates complex human decision-making processes. Conversations end when the user simulator sends an end token.

\section{UserLM with Reasoning}
We propose UserLM-R1, a user simulator with reasoning capabilities, as shown in Figure \ref{fw}. First, we construct generalizable user profiles by separating user personas into static roles and dynamic concerns (Sec. \ref{profile}). Then, we propose a goal-driven decision-making strategy that performs reasoning prior to response generation (Sec. \ref{reasoning}), bootstrap analysis abilities of the model using supervised fine-tuning (Sec. \ref{sft}), and further strengthen its behavioral fidelity via reinforcement learning with multi-reward (Sec. \ref{rl}).

\subsection{General User Profile Construction}
\label{profile}
To advance general user simulation across various application scenarios, we require diverse and comprehensive user personas. Current methods extract roles from open-source resources like Wikipedia or TV series. However, these historical or fictional characters often diverge significantly from real-life user behavior and suffer from high homogenization, lacking the variability for robust general simulation, while these generated profiles are typically scenario-specific, which severely hinders their transferability and scalability across different applications. To address these challenges, we propose to extract personas from social media and real business scenarios. Further, we decouple the user profile into static roles and dynamic concerns, enhancing the transferability and extensibility of our user personas. As shown in Table \ref{tab:profile_schema}, the static profile focuses on task-consistent attributes, including background information, personality traits, expression style, and life scenarios, which remain static across different task scenarios. In contrast, the dynamic profile includes scenario memory, target lists, decision policies, and state change transitions that evolve across different task scenarios.

Specifically, we conduct profile enhancement on AlignX \cite{DBLP:journals/corr/abs-2503-15463}. The original user preference samples extracted from Reddit contain a 90-dimensional preference vector and the demographic information for each user. We first convert the vector into coherent narrative descriptions based on the feature space. Then, by using detailed instructions and optional attribute lists, we leverage an advanced large language model to generate each user character attribute in the static profile with human verification. For the dynamic profile, it is generated on demand for each task agent according to its specific business standard operating procedure (SOP) and evolves continuously as the conversation progresses. Accordingly, for each user-agent pair, we construct scenario-specific memory aligned with the SOP and grounded in the static user profile. Detailed explanations and corresponding prompts can be found in Appendix \ref{profile_construction}.

\subsection{Goal-Driven Decision-Making Strategy}
\label{reasoning}
Typically, the user model is provided with the user profile $\bm{p}_i$ in the system prompt and generates responses based solely on conversation history. At each turn $j$,  
\begin{equation}
    \bm{u}_{ij} = \pi_{\theta}(\bm{p}_i, \bm{c}_{ij}),
\end{equation}
where $\pi_{\theta}$ is the user model, and  $\bm{c}_{ij}$ is the conversation context. However, this formulation lacks an explicit reasoning process, making it fail to capture the cognitive nuances of authentic human interaction, where individuals typically analyze the current dialogue state and reflect on underlying concerns before planning their subsequent response. To bridge this gap, we propose to reconstruct the model interaction process into the system encompassing explicit cognitive reasoning and dynamic state evolution. Instead of relying solely on the static user profile $\bm{p}_i^s$, we introduce the dynamic user profile $\bm{p}_{ij}^d$ to capture the transient psychological status of users at turn $j$, e.g., emotional fluctuations and shifts in concerns. Under this paradigm, response generation transforms from a direct mapping into a sequential decision-making process. Conditioned on the dialogue context and the current user profile, the user model first generates an explicit chain of thought $\bm{r}_{ij}$ to analyze the ongoing situation and plan the subsequent action. It then produces an explicit decision policy and state value.

 Specifically, the decision rationale in the reasoning process is decomposed into five subtasks: recognizing agent intent, organizing user concerns, planning next action, updating state values, and refining the response tone. In addition, we introduce a set of operational guidelines, including touched concerns and issues, topic management rules, next step planning, hang-up conditions, and restrictions on prohibited behaviors and expressions. For state value updates, we model trust, emotion, patience, and participation as key factors by defining discrete levels for each state and assign corresponding values that guide subsequent behaviors and responses. Tracking these states helps maintain consistency and coherence. For the first turn, the model generates a target list comprising primary and secondary concerns that remains consistent throughout the session.

\subsection{Bootstrapping via Supervised Fine-Tuning}
\label{sft}
To empower the user model with reasoning capability to simulate diverse users, we train the user simulator on user reasoning text and responses from the generated conversation at the utterance level conditioned on user profiles. We use the standard supervised fine-tuning objective:
\begin{equation}
\mathop{\min}\limits_{\pi_{\theta}} \sum_{i, j} -\log \pi_{\theta}(\bm{r}_{ij}, \bm{u}_{ij} | I, \bm{p}_{i}^s, \bm{p}_{ij}^d, \bm{c}_{ij}),
\end{equation}
where $I$ denotes the simulation instruction, $\bm{c}_{ij}$ is the conversation history, and $\bm{p}_{i}^s$ and $\bm{p}_{ij}^d$ represent the static profile and $j$-th turn dynamic profile for the $i$-th user, respectively. The expected generated content $\bm{r}_{ij}$ and $ \bm{u}_{ij}$ correspond to the reasoning rationales and responses for the $i$-th user at the $j$-th turn. We extract updated dynamic profiles from the generated responses and incorporate them into the subsequent dialogue context, thereby explicitly maintaining the consistency of behavior of the user simulator across turns. Through SFT, the user model learns to perform inference, evaluate arguments and then draw logically grounded responses based on available information.

\begin{table*}[t]
\centering
\begin{tabular}{lcccccccccc}
\toprule
\multirow{2}{*}{\textbf{Method}} & \multicolumn{4}{c}{\textbf{Session-Level}}                                          & \multicolumn{6}{c}{\textbf{Turn-Level}}                                              \\ \cmidrule(r){2-5} \cmidrule(r){6-11}
                                 & \textbf{Role} & \textbf{Int.} & \textbf{Goal} & \textbf{Total} & \textbf{Rob. $\downarrow$} & \textbf{CoT} & \textbf{Stra.} & \textbf{Pers.} & \textbf{Con.} & \textbf{Total} \\ \midrule
\rowcolor{gray!20}
\multicolumn{11}{c}{\textit{\textbf{General Baselines}}}  \\ 
Qwen3-8B                                            & 85.57                    & 29.85                    & 34.21                    & 58.80    & 86.40 & -     & 38.30 & 41.55 & -     & -     \\
w/ think                                            & 85.24 & 30.25 & 37.88 & 59.65                    & 31.40                & 39.86                & 48.73                & 57.73                & 76.64                & 48.47                \\
Qwen3-32B                                           & 87.67                    & 38.92                    & 37.96                    & 63.06                     & 66.40 & -     & 42.98 & 47.64 & -     & -     \\
w/ think                                             & 89.29                    & 37.71                    & 41.46                    & 64.44                     & 12.30                & 41.82                & 52.86                & 62.39                & 80.14                & 53.38                \\
DeepSeek-R1                                          & \underline{94.29}                    & \textbf{58.75}                    & \underline{46.92}                    & \textbf{73.56}                     & \textbf{4.10}                 & 45.57                & 57.15                & 66.92                & 82.97                & 58.01                \\
Doubao-Seed-1.6                                     & 84.58                    & 47.54                    & 39.71                    & 64.10            & 9.50                 & 41.91                & 52.64                & 63.80                & 80.73                & 53.75                \\
Gemini-2.5-Flash                                     & 92.00                    & 54.87                    & 40.79                    & 69.92          & 6.40                 & 49.09                & 59.11                & 69.64                & \underline{83.50}                & 60.17                \\ 
 \midrule
\rowcolor{gray!20}
\multicolumn{11}{c}{\textit{\textbf{User Simulator Baselines}}}                    \\
CharacterGLM                                         & 45.58                    & 13.87                     & 14.50                    & 29.88                     & 95.00                & -                    & 33.97                & 37.61                & -                    & -                    \\
Xingchen                                             & 86.09                    & \underline{56.34}                    & 44.70                    & 68.31                     & 78.60                & -                    & 38.30                & 43.05                & -                    & -                    \\  \midrule
\rowcolor{gray!20}
\multicolumn{11}{c}{\textit{\textbf{Ours}}} \\
UserLM-R1-8B                                         & 92.25                    & 32.71                    & 45.96                    & 65.79                     & 14.10                & \underline{55.52}                & \underline{59.98}                & \underline{69.66}                & 82.27                & \underline{61.55}                \\
UserLM-R1-32B                                        & \textbf{95.21}                    & 46.08                    & \textbf{57.46}                    & \underline{73.49}                     & \underline{4.50}                 & \textbf{73.39}                & \textbf{74.52}                & \textbf{80.50}                & \textbf{92.55}               & \textbf{76.56}               
           \\ \bottomrule
\end{tabular}
\caption{Experimental results at session and turn levels. The best and second-best performing models are indicated in bold and underlined, respectively. $\downarrow$ denotes that a lower value is better.}
\label{main_res}
\end{table*}

\subsection{Generalizing with Multi-Reward Reinforcement Learning}
\label{rl}
The supervised fine-tuning stage minimizes the discrepancy between model outputs and training samples but inherently constrains the model ability to generalize beyond the distribution of the annotated data. However, in open-ended dialogues, viable responses are often diverse and non-unique. To address this limitation, we introduce reinforcement learning  to facilitate the model exploring more logical, diverse, or strategically intelligent reasoning trajectories than those in the original dataset. Specifically, we design a multi-reward scheme that incorporates both rule-based and rubric-based rewards, providing dense and fine-grained supervision signals. Given an $n$-turn dialogue  $\bm{d}_i = \{(\bm{a}_{i, 1}, \bm{r}_{i, 1}, \bm{u}_{i, 1}), ..., (\bm{a}_{i, n}, \bm{r}_{i, n}, \bm{u}_{i, n})\}$, we calculate the following rewards.

\noindent \textbf{Rule-Based Rewards.}  To guide the model toward producing the correct reasoning format, we first use the format-checking rule-based rewards $R_{\text{rule}}$. Specifically, we define constraints for the rationales and responses, i.e., verify the \texttt{<think></think>} and \texttt{<answer></answer>} for each turn. Furthermore, to encourage detailed reasoning, we penalize thinking processes that are of insufficient length. Regarding the response content, we mandate the inclusion of specific fields defined in the dynamic profiles, and the absence of these fields results in a score deduction. Finally, the score is clipped to the range $[0, 1]$.

\noindent \textbf{Rubric-Based Rewards.} We design rewards based on the rubric set $\mathcal{C}_{\text{rubric}}$, which covers response consistency, reasoning quality, alignment of reasoning and response, and strategic capability. Specifically, the response consistency reward evaluates the reply fidelity to the character profile, stylistic coherence, and oral naturalness, thereby ensuring the generation of authentic, human-like dialogue strictly aligned with the target profiles. The reasoning quality reward comprehensively evaluates goal management precision, state transition validity, and reasoning depth by penalizing template-based redundancy, thereby fostering rigorous, multi-step strategic reasoning. The alignment reward assesses the faithful execution of core strategies formulated in the chain of thought. The strategic capability reward incentivizes accurate trap identification, state alignment, and, crucially, the active seizing of conversational initiative through logical counterattacks, empowering the model to robustly dominate deceptive scenarios. Given a generated answer with rationales and responses $(\bm{r}_{i, k}, \bm{u}_{i, k})$, the rubric reward is defined as follows:

\begin{equation}
     R_{\text{rubric}} = \frac{1}{|\mathcal{C}_{\text{rubric}}|}\sum_{\mathbb{C}_m \in \mathcal{C}_{\text{t}}} \mathbb{C}_m(\bm{r}_{i,k}, \bm{u}_{i,k}),
\end{equation}
where $\mathbb{C}_m(\cdot, \cdot)$ represents reward function. In contrast to SFT, which teaches the model to generate responses through next-token prediction, during the reinforcement learning stage, we employ the GRPO algorithm with rule-based and rubric-based rewards as guiding signals to instruct the model on how to achieve its objectives.

\section{Experiments}

\subsection{Experimental Setup}

\noindent\textbf{Datasets  }  We extract 90k user profiles from the large-scale personalized preference dataset AlignX \cite{DBLP:journals/corr/abs-2503-15463}, following the procedure described in Sec. \ref{profile}. For agent SOPs, we refine the task SOPs covering six real business scenarios from VoiceAgentEval \cite{DBLP:journals/corr/abs-2510-21244} and further augment them, resulting in 450 distinct SOPs. We then construct 1,440 tasks by pairing user profiles with SOPs. Further, to rigorously evaluate the logical reasoning and strategic stability of user simulators, we argue that standard interaction scenarios are insufficient, as they often assume a cooperative agent. Therefore, we construct a challenging adversarial dataset consisting of 220 samples with 11 adversarial strategies. In this setting, the agent actively employs aggressive persuasion strategies, such as cognitive traps, false urgency, and sunk cost fallacies, to induce the user simulator into abandoning its original goals or accepting sub-optimal deals. Details are in Appendix \ref{adv_dataset}.

\noindent\textbf{Baselines  }  We compare our method, UserLM-R1, with open-source models, advanced commercial models, and dedicated user simulator baselines on our benchmarks. Open-source baselines include Qwen3-8B and Qwen3-32B \cite{DBLP:journals/corr/abs-2505-09388}, evaluated both with and without chain-of-thought generation. Advanced commercial models include Gemini-2.5-Flash \cite{DBLP:journals/corr/abs-2507-06261}, Doubao-Seed-1.6 \cite{seed1.6} and DeepSeek-R1 \cite{DBLP:journals/corr/abs-2501-12948}. User simulator baselines include CharacterGLM \cite{DBLP:conf/emnlp/Zhou0WWSYHKBPYX24}, the role-play user model derived from ChatGLM,  and Tongyi Xingchen \cite{xingchen}, a close-sourced character-based dialogue agent developed by Alibaba Cloud.

\noindent\textbf{Implementation Details }
We use the Qwen-3 series in two sizes, i.e., 8B and 32B, as base LLMs. For supervised fine-tuning, we train all models for three epochs with a learning rate of $5\times 10^{-5}$ , and a generated sequence length of $3,076$. For reinforcement learning, we use a learning rate of $1\times 10^{-6}$. During the rollout phase, we generate $8$ responses per sample with a batch size of $64$. The coefficient for Kullback-Leibler divergence is set to $0.001$. During evaluation, we use GPT-4o \cite{DBLP:journals/corr/abs-2410-21276} as the LLM judger.

\begin{table*}[t]
\centering
\begin{tabular}{lcccccccccc}
\toprule
\multirow{2}{*}{\textbf{Method}} & \multicolumn{4}{c}{\textbf{Session-Level}}                                          & \multicolumn{6}{c}{\textbf{Turn-Level}}                                              \\ \cmidrule(r){2-5} \cmidrule(r){6-11}
                                 & \textbf{Role} & \textbf{Int.} & \textbf{Goal} & \textbf{Total} & \textbf{Rob. $\downarrow$} & \textbf{CoT} & \textbf{Stra.} & \textbf{Pers.} & \textbf{Con.} & \textbf{Total} \\ \midrule
Qwen3-8B           & 85.57 & 29.85 & 34.21 & 58.80 & 86.40 & -     & 38.30 & 41.55 & -     & -     \\
\; w/ think            & 85.24 & 30.25 & 37.88 & 59.65 & 31.40 & 39.86 & 48.73 & 57.73 & 76.64 & 48.47 \\
\; w/ Goal. Strategy & 86.33 & 31.41 & 38.02 & 60.22 & 27.70 & 37.55 & 53.45 & 66.52 & 73.18 & 51.39 \\
\; w/ SFT              & 90.29 & 32.00 & 39.42 & 63.00 & 19.50 & 51.91 & 56.86 & 66.84 & 79.36 & 58.03 \\
\; w/ RL         & \textbf{92.25} & \textbf{32.71} & \textbf{45.96} & \textbf{65.79} & \textbf{14.10} & \textbf{55.52} & \textbf{59.98} & \textbf{69.66} & \textbf{82.27} & \textbf{61.55}
        \\ \bottomrule
\end{tabular}
\caption{Ablation study results.}
\label{ablation_res}
\end{table*}

\subsection{Evaluation Metrics}
We evaluate user simulators from both the session-level and turn-level perspectives. For session-level evaluation, we sample 120 tasks from the constructed dataset as the test set and assess the entire dialogues. In contrast, for turn-level evaluation, we focus on the performance of the models when confronted with challenging adversarial agent turns. Specifically, the session-level evaluation includes three metrics: Role authenticity (Role), Interaction performance (Int.), and Goal progress (Goal). For turn-level evaluation, we assess five dimensions: Robotic tone (Rob.), CoT effectiveness (CoT), Game-theoretic strategy (Stra.), Persona fidelity (Pers.), and Thought-response consistency (Con.). Detailed definitions of the evaluation metrics can be found in Appendix \ref{eval_metric}.

\subsection{Main Results}

Table \ref{main_res} reports the session-level and turn-level results for general baselines, role-play expertise baselines and our method. For session-level results, we can observe that open-source general baselines still perform worse in interaction strategy and goal progress. Closed-source general baselines achieve promising interaction performance, outperforming other general models. Among the role-playing expertise baselines, the open-source models exhibit significantly lower overall performance compared to other models, while the proprietary Xingchen performs better. Our model, UserLM-R1-32B, demonstrates a substantial performance gain over conventional user simulator baselines, such as CharacterGLM and Xingchen. Notably, on the goal progress metric, UserLM-R1-32B outperforms all baselines, highlighting its superior conversational proactivity and strategic maneuvering. This advantage enables our model to guide dialogues more effectively through realistic and reliable user simulation.

In the turn-level evaluation, where more challenging adversarial samples are involved, all baselines experience significant drops, particularly traditional user simulators. From the first four lines, we observe that the integration of the reasoning mechanism effectively reduces the robotic tone and enhances both the game-theoretic ability and persona fidelity. UserLM-R1-32B achieves the highest performance on reasoning content and strategy, indicating that the model not only identifies conversational traps but also proactively seizes the initiative from the agent. Additionally, the persona fidelity and consistency scores confirm that the model maintains the human-centric affective style and effective reasoning. The UserLM-R1-8B variant also delivers impressive results, further validating the effectiveness of our approach. Detailed results are in Appendix \ref{adv_turn_eval}.

\begin{figure}[t]
	\centering
	\subfigure[Quality Analysis]{
		\includegraphics[height=3.3cm,width=3.6cm]{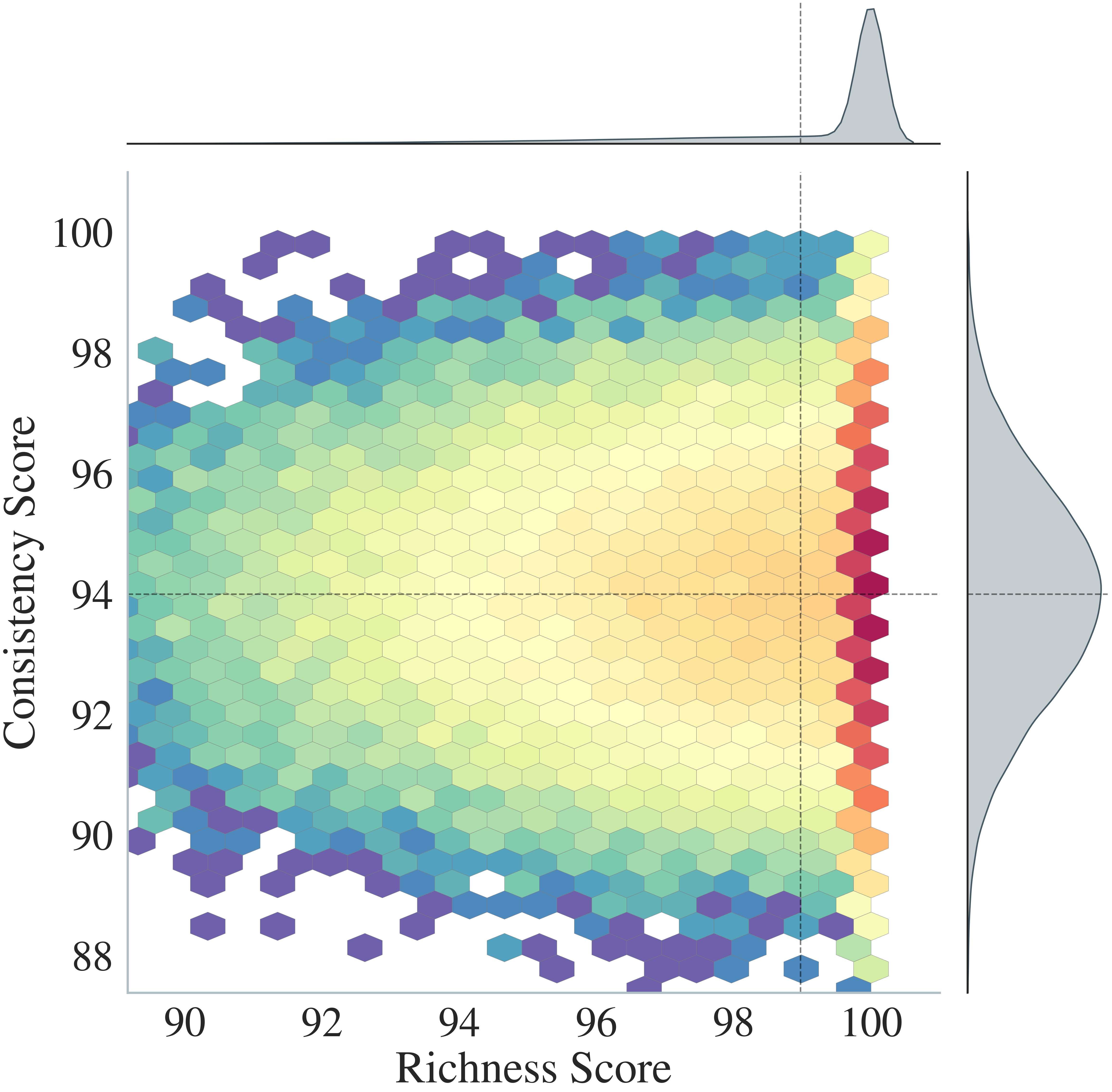}} 
	\subfigure[Expression Style]{
		\includegraphics[height=3.3cm,width=3.6cm]{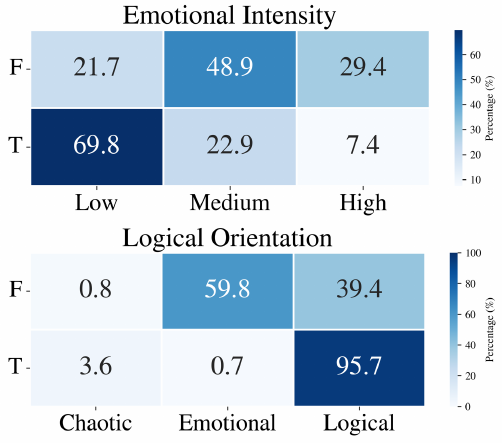}
	}
	\caption{Quality and expression style analysis of the constructed user profiles.}
	\label{quality_analysis}
\end{figure}

\subsection{Ablation Study}

We conduct ablation studies using UserLM-R1-8B to assess the relative contributions of each component of our method in Table \ref{ablation_res}. We incrementally incorporate the vanilla reasoning process, the goal-driven strategy, supervised fine-tuning, and reinforcement learning. As illustrated in Table \ref{ablation_res}, introducing the reasoning process yields performance gains, while integrating the goal-driven strategy leads to substantially larger improvements on both session-level and turn-level metrics, validating its effectiveness. For supervised fine-tuning, the results indicate that distilling examples consistently produces stronger performance. Finally, reinforcement learning further enhances overall results because rewards encourage the model to explore more coherent and varied trajectories beyond those in the annotated dataset.

\section{Further Analysis}

\subsection{Profile Analysis}

We conduct a comprehensive user profile evaluation for assessing their quality and representativeness, which are essential for modeling real-world population characteristics. From Figure \ref{quality_analysis}(a), we can see that user profiles are predominantly distributed in regions with high consistency and richness. Figure \ref{quality_analysis}(b) visualizes the relationship between emotional intensity, logical orientation, and MBTI personality traits. We can see that Thinkers (T) tend to display lower emotional intensity and stronger logical orientation, while Feelers (F) are characterized by moderate to high emotional intensity and affect-oriented expression, indicating that the assigned MBTI personality attributes are well aligned with the corresponding expression styles. Further basic information and multi-dimensional correlation analysis can be found in Appendix \ref{ap_profile_eva}.

\begin{table}[t]
\centering
\begin{tabular}{lcc}
\toprule
\multirow{2}{*}{Method} & \multicolumn{2}{c}{Metrics }                                                                                                                              \\ \cmidrule{2-3} 
                          & \begin{tabular}[c]{@{}c@{}}Session\\ ($\kappa=0.63$)\end{tabular} & \begin{tabular}[c]{@{}c@{}}Turn\\ ($\kappa=0.72$)\end{tabular} \\ \midrule
Gemini-2.5-Flash                    & 86/22/12                                                       & 142/40/38                                                                                                       \\
DeepSeek-R1                   &  44/28/48                                                       & 168/20/42                                                                                                                    \\
Ours (w/o RL)             &   55/32/33                                                 &   67/111/42                                                                                                           \\ \bottomrule
\end{tabular}
\caption{Human evaluation results of UserLM-R1-32B win/tie/loss rates. $\kappa$ denotes the kappa coefficient.}
\label{human_res}
\end{table}

\subsection{Human Evaluation}
We recruit three evaluators to assess the quality of sessions and turns from the test set and the adversarial set, respectively.
Table \ref{human_res} reports the win/tie/loss rates of UserLM-R1-32B against other models, as well as the with-in group kappa coefficients. The results show that UserLM-R1 consistently outperforms Gemini-2.5-Flash and DeepSeek-R1 on two datasets, demonstrating the effectiveness of our approach. We also evaluate the variant of UserLM-R1 without reinforcement learning and observe performance declines, further confirming that RL encourages the model to learn more diverse and robust behavioral policies. We further conduct case studies in Appendix \ref{case_study}.

\begin{figure}[t]
	\centering
	\includegraphics[width=1\linewidth]{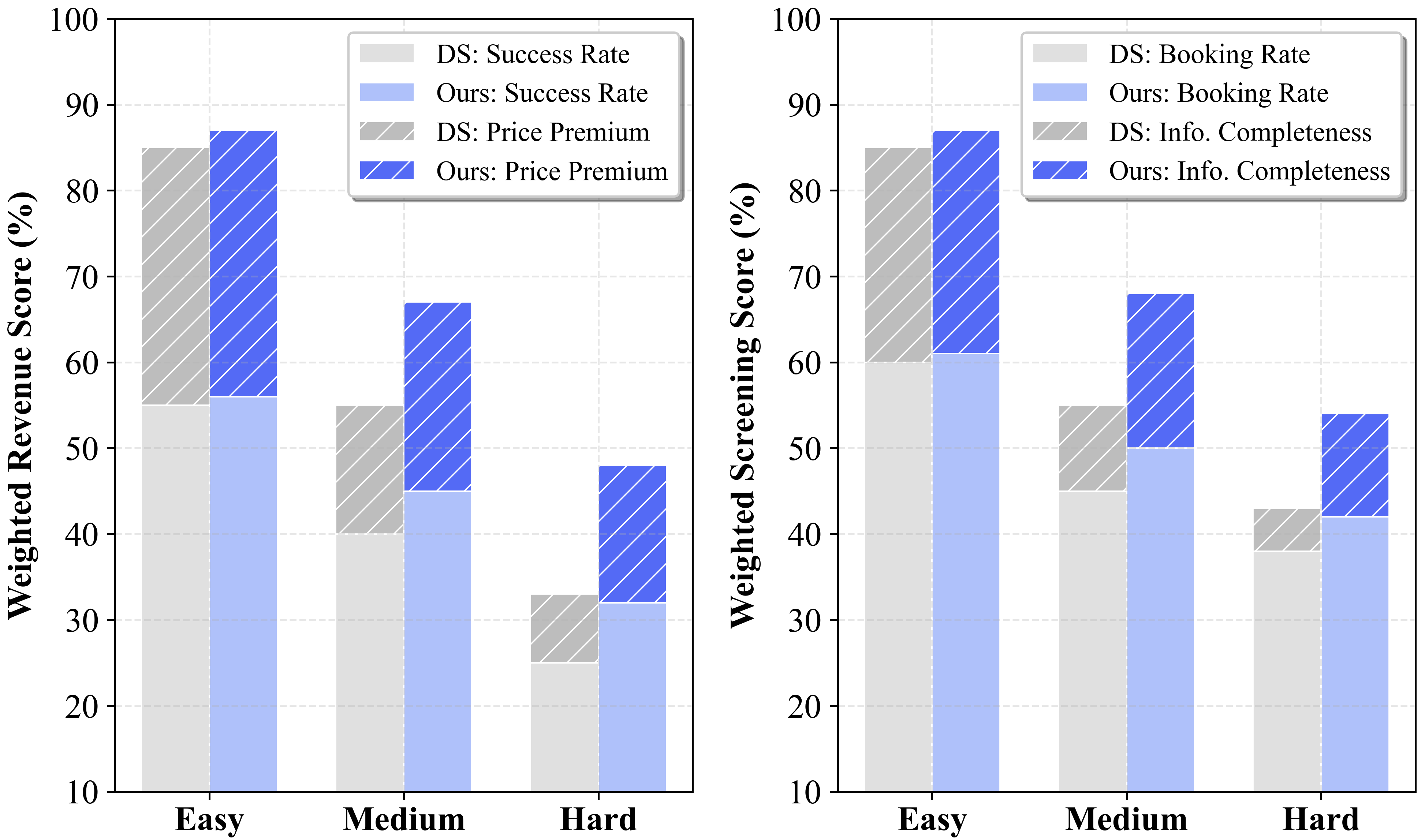}
	\caption{Evaluation results of retail and hiring agents trained with DeepSeek-R1 and UserLM-R1.}
	\label{agent_res}
\end{figure}

\subsection{Effectiveness in RL Agent Training}
One important application of the user simulator is its ability to interact with task agents to generate rollout samples in RL training. We deploy UserLM-R1 in developing industrial retail and hiring agents, i.e., sales negotiation and candidate screening. During RL, we compare DeepSeek-R1 and UserLM-R1 as user simulators for interactive training. We categorize the targeted real users into three groups according to their domain proficiency and degree of critical standards. As shown in Figure \ref{agent_res}, the agent trained with UserLM-R1 achieves consistently better performance, particularly in scenarios involving budget-constrained and assertive users in sales, as well as cautious and exacting experts in hiring. This improvement demonstrates the superior negotiation capability of UserLM-R1 and underscores its effectiveness in real-world industrial applications.

\section{Conclusion}
This paper introduces a novel reasoning user simulation framework called UserLM-R1. We first propose general user profiles with both static user roles and dynamic scenario-specific goals, making our profiles scalable and generalizable. Additionally, we propose the goal-driven reasoning strategy covering the decision trajectories and state changes before generating responses, and we further
enhance reasoning via multi-reward reinforcement learning. Explicitly tracking mental states and key context enhances realism and long-term consistency, preventing agents from manipulating the simulator via shortcuts. Experimental results demonstrate that UserLM-R1 significantly improves performance in achieving profile adherence, fostering proactive conversational engagement, and demonstrating higher human-likeness.

\section*{Limitations}

While the adoption of dynamic profiles in our framework mitigates domain generalization issues, there remains a gap in replicating real human interaction. Current profiles lack the depth of lifelong episodic and semantic memories that fundamentally shape human cognition and decision-making. Consequently, developing mechanisms for simulators to learn from, organize, and retrieve long-term experiences from past dialogues remains an open area for future research. Additionally, we focus on Chinese samples in this paper. Future research may broaden this scope to multilingual settings and investigate how cultural differences influence user behavioral patterns.

\section*{Ethical Statement}

Regarding data privacy, our constructed user profile pool is free from real user personal information leakage. Profile attributes, such as names, occupations, and demographics, are synthetically generated by large language models and do not correspond to real individuals. For potential risks, although we have implemented prompt constraints and conducted ethical checks to mitigate these issues, we acknowledge that generative models may inherently produce offensive or biased content. In the future, careful alignment for safety will minimize these risks. For the use of large language models, we use large language models solely to enhance the grammar of our drafts.  Importantly, all conceptual contributions, including the design of research questions and the development of methods, are conceived and carried out entirely by the authors.

\bibliography{custom}

@article{DBLP:journals/corr/abs-2501-04410,
  author       = {Krisztian Balog and
                  ChengXiang Zhai},
  title        = {User Simulation in the Era of Generative {AI:} User Modeling, Synthetic
                  Data Generation, and System Evaluation},
  journal      = {CoRR},
  volume       = {abs/2501.04410},
  year         = {2025},
  url          = {https://doi.org/10.48550/arXiv.2501.04410},
  eprinttype    = {arXiv}
}

@article{DBLP:journals/corr/abs-2406-12045,
  author       = {Shunyu Yao and
                  Noah Shinn and
                  Pedram Razavi and
                  Karthik Narasimhan},
  title        = {{\(\tau\)}-bench: {A} Benchmark for Tool-Agent-User Interaction in
                  Real-World Domains},
  journal      = {CoRR},
  volume       = {abs/2406.12045},
  year         = {2024},
  url          = {https://doi.org/10.48550/arXiv.2406.12045},
  doi          = {10.48550/ARXIV.2406.12045},
  eprinttype    = {arXiv}
}

@article{argyle2023out,
  title={Out of one, many: Using language models to simulate human samples},
  author={Argyle, Lisa P and Busby, Ethan C and Fulda, Nancy and Gubler, Joshua R and Rytting, Christopher and Wingate, David},
  journal={Political Analysis},
  volume={31},
  number={3},
  pages={337--351},
  year={2023}
}

@inproceedings{DBLP:conf/uist/ParkOCMLB23,
  author       = {Joon Sung Park and
                  Joseph C. O'Brien and
                  Carrie Jun Cai and
                  Meredith Ringel Morris and
                  Percy Liang and
                  Michael S. Bernstein},
  title        = {Generative Agents: Interactive Simulacra of Human Behavior},
  booktitle    = {Annual {ACM} Symposium on User Interface Software
                  and Technology (UIST)},
  pages        = {2:1--2:22},
  publisher    = {{ACM}},
  year         = {2023},
  url          = {https://doi.org/10.1145/3586183.3606763}
}

@inproceedings{DBLP:conf/emnlp/Zhou0WWSYHKBPYX24,
  author       = {Jinfeng Zhou and
                  Zhuang Chen and
                  Dazhen Wan and
                  Bosi Wen and
                  Yi Song and
                  Jifan Yu and
                  Yongkang Huang and
                  Pei Ke and
                  Guanqun Bi and
                  Libiao Peng and
                  Jiaming Yang and
                  Xiyao Xiao and
                  Sahand Sabour and
                  Xiaohan Zhang and
                  Wenjing Hou and
                  Yijia Zhang and
                  Yuxiao Dong and
                  Hongning Wang and
                  Jie Tang and
                  Minlie Huang},
  title        = {CharacterGLM: Customizing Social Characters with Large Language Models},
  booktitle    = {Conference on Empirical Methods in Natural
                  Language Processing (EMNLP)},
  pages        = {1457--1476},
  year         = {2024},
  url          = {https://doi.org/10.18653/v1/2024.emnlp-industry.107}
}

@article{DBLP:journals/corr/abs-2401-09432,
  author       = {Meiling Tao and
                  Xuechen Liang and
                  Tianyu Shi and
                  Lei Yu and
                  Yiting Xie},
  title        = {RoleCraft-GLM: Advancing Personalized Role-Playing in Large Language
                  Models},
  journal      = {CoRR},
  volume       = {abs/2401.09432},
  year         = {2024},
  url          = {https://doi.org/10.48550/arXiv.2401.09432},
  doi          = {10.48550/ARXIV.2401.09432},
  eprinttype    = {arXiv}
}

@inproceedings{DBLP:conf/acl/WangPQLZWGGN00024,
  author       = {Noah Wang and
                  Zhongyuan Peng and
                  Haoran Que and
                  Jiaheng Liu and
                  Wangchunshu Zhou and
                  Yuhan Wu and
                  Hongcheng Guo and
                  Ruitong Gan and
                  Zehao Ni and
                  Jian Yang and
                  Man Zhang and
                  Zhaoxiang Zhang and
                  Wanli Ouyang and
                  Ke Xu and
                  Wenhao Huang and
                  Jie Fu and
                  Junran Peng},
  title        = {RoleLLM: Benchmarking, Eliciting, and Enhancing Role-Playing Abilities
                  of Large Language Models},
  booktitle    = {Findings of the Association for Computational Linguistics, {ACL}},
  pages        = {14743--14777},
  year         = {2024},
  url          = {https://doi.org/10.18653/v1/2024.findings-acl.878}
}

@inproceedings{DBLP:conf/acl/Liu0YWM0WW23,
  author       = {Yajiao Liu and
                  Xin Jiang and
                  Yichun Yin and
                  Yasheng Wang and
                  Fei Mi and
                  Qun Liu and
                  Xiang Wan and
                  Benyou Wang},
  title        = {One Cannot Stand for Everyone! Leveraging Multiple User Simulators
                  to train Task-oriented Dialogue Systems},
  booktitle    = {Proceedings of the 61st Annual Meeting of the Association for Computational
                  Linguistics (ACL) },
  pages        = {1--21},
  year         = {2023},
  url          = {https://doi.org/10.18653/v1/2023.acl-long.1}
}

@article{DBLP:journals/corr/abs-2504-03601,
  author       = {Akshara Prabhakar and
                  Zuxin Liu and
                  Ming Zhu and
                  Jianguo Zhang and
                  Tulika Awalgaonkar and
                  Shiyu Wang and
                  Zhiwei Liu and
                  Haolin Chen and
                  Thai Hoang and
                  Juan Carlos Niebles and
                  Shelby Heinecke and
                  Weiran Yao and
                  Huan Wang and
                  Silvio Savarese and
                  Caiming Xiong},
  title        = {APIGen-MT: Agentic Pipeline for Multi-Turn Data Generation via Simulated
                  Agent-Human Interplay},
  journal      = {CoRR},
  volume       = {abs/2504.03601},
  year         = {2025},
  url          = {https://doi.org/10.48550/arXiv.2504.03601},
  doi          = {10.48550/ARXIV.2504.03601},
  eprinttype    = {arXiv}
}

@article{DBLP:journals/corr/abs-2509-19736,
  author       = {Cheng Qian and
                  Zuxin Liu and
                  Akshara Prabhakar and
                  Jielin Qiu and
                  Zhiwei Liu and
                  Haolin Chen and
                  Shirley Kokane and
                  Heng Ji and
                  Weiran Yao and
                  Shelby Heinecke and
                  Silvio Savarese and
                  Caiming Xiong and
                  Huan Wang},
  title        = {UserRL: Training Interactive User-Centric Agent via Reinforcement
                  Learning},
  journal      = {CoRR},
  volume       = {abs/2509.19736},
  year         = {2025},
  url          = {https://doi.org/10.48550/arXiv.2509.19736},
  doi          = {10.48550/ARXIV.2509.19736},
  eprinttype    = {arXiv}
}

@article{DBLP:journals/corr/abs-2309-13233,
  author       = {Sam Davidson and
                  Salvatore Romeo and
                  Raphael Shu and
                  James Gung and
                  Arshit Gupta and
                  Saab Mansour and
                  Yi Zhang},
  title        = {User Simulation with Large Language Models for Evaluating Task-Oriented
                  Dialogue},
  journal      = {CoRR},
  volume       = {abs/2309.13233},
  year         = {2023},
  url          = {https://doi.org/10.48550/arXiv.2309.13233},
  doi          = {10.48550/ARXIV.2309.13233},
  eprinttype    = {arXiv}
}

@article{DBLP:journals/tois/SunGZRCRR24,
  author       = {Weiwei Sun and
                  Shuyu Guo and
                  Shuo Zhang and
                  Pengjie Ren and
                  Zhumin Chen and
                  Maarten de Rijke and
                  Zhaochun Ren},
  title        = {Metaphorical User Simulators for Evaluating Task-oriented Dialogue
                  Systems},
  journal      = {{ACM} Trans. Inf. Syst.},
  volume       = {42},
  number       = {1},
  pages        = {17:1--17:29},
  year         = {2024},
  url          = {https://doi.org/10.1145/3596510}
}

@inproceedings{DBLP:conf/naacl/LuHZANBMMLYWP25,
  author       = {Jiarui Lu and
                  Thomas Holleis and
                  Yizhe Zhang and
                  Bernhard Aumayer and
                  Feng Nan and
                  Haoping Bai and
                  Shuang Ma and
                  Shen Ma and
                  Mengyu Li and
                  Guoli Yin and
                  Zirui Wang and
                  Ruoming Pang},
  title        = {ToolSandbox: {A} Stateful, Conversational, Interactive Evaluation
                  Benchmark for {LLM} Tool Use Capabilities},
  booktitle    = {Findings of the Association for Computational Linguistics: {NAACL}},
  pages        = {1160--1183},
  year         = {2025},
  url          = {https://doi.org/10.18653/v1/2025.findings-naacl.65}
}

@inproceedings{DBLP:conf/acl/WangL000025,
  author       = {Kuang Wang and
                  Xianfei Li and
                  Shenghao Yang and
                  Li Zhou and
                  Feng Jiang and
                  Haizhou Li},
  title        = {Know You First and Be You Better: Modeling Human-Like User Simulators
                  via Implicit Profiles},
  booktitle    = {Annual Meeting of the Association for Computational
                  Linguistics (ACL)},
  pages        = {21082--21107},
  year         = {2025},
  url          = {https://aclanthology.org/2025.acl-long.1025/}
}

@inproceedings{DBLP:conf/acl/YuYWZQ25,
  author       = {Yeyong Yu and
                  Runsheng Yu and
                  Haojie Wei and
                  Zhanqiu Zhang and
                  Quan Qian},
  title        = {Beyond Dialogue: {A} Profile-Dialogue Alignment Framework Towards
                  General Role-Playing Language Model},
  booktitle    = {Annual Meeting of the Association for Computational
                  Linguistics (ACL) },
  pages        = {11992--12022},
  year         = {2025},
  url          = {https://aclanthology.org/2025.acl-long.586/}
}

@article{DBLP:journals/corr/abs-2402-13374,
  author       = {Ivan Sekulic and
                  Silvia Terragni and
                  Victor Guimar{\~{a}}es and
                  Nghia Khau and
                  Bruna Guedes and
                  Modestas Filipavicius and
                  Andr{\'{e}} Ferreira Manso and
                  Roland Mathis},
  title        = {Reliable LLM-based User Simulator for Task-Oriented Dialogue Systems},
  journal      = {CoRR},
  volume       = {abs/2402.13374},
  year         = {2024},
  url          = {https://doi.org/10.48550/arXiv.2402.13374},
  doi          = {10.48550/ARXIV.2402.13374},
  eprinttype    = {arXiv}
}

@article{DBLP:journals/corr/abs-2511-00222,
  author       = {Marwa Abdulhai and
                  Ryan Cheng and
                  Donovan Clay and
                  Tim Althoff and
                  Sergey Levine and
                  Natasha Jaques},
  title        = {Consistently Simulating Human Personas with Multi-Turn Reinforcement
                  Learning},
  journal      = {CoRR},
  volume       = {abs/2511.00222},
  year         = {2025},
  url          = {https://doi.org/10.48550/arXiv.2511.00222},
  doi          = {10.48550/ARXIV.2511.00222},
  eprinttype    = {arXiv}
}

@article{DBLP:journals/corr/abs-2510-06552,
  author       = {Tarek Naous and
                  Philippe Laban and
                  Wei Xu and
                  Jennifer Neville},
  title        = {Flipping the Dialogue: Training and Evaluating User Language Models},
  journal      = {CoRR},
  volume       = {abs/2510.06552},
  year         = {2025},
  url          = {https://doi.org/10.48550/arXiv.2510.06552},
  doi          = {10.48550/ARXIV.2510.06552},
  eprinttype    = {arXiv}
}

@article{DBLP:journals/corr/abs-2507-20152,
  author       = {Shuhaib Mehri and
                  Xiaocheng Yang and
                  Takyoung Kim and
                  Gokhan Tur and
                  Shikib Mehri and
                  Dilek Hakkani{-}T{\"{u}}r},
  title        = {Goal Alignment in LLM-Based User Simulators for Conversational {AI}},
  journal      = {CoRR},
  volume       = {abs/2507.20152},
  year         = {2025},
  url          = {https://doi.org/10.48550/arXiv.2507.20152},
  doi          = {10.48550/ARXIV.2507.20152},
  eprinttype    = {arXiv}
}

@inproceedings{DBLP:conf/www/ChenDZFZT0ZD25,
  author       = {Luyu Chen and
                  Quanyu Dai and
                  Zeyu Zhang and
                  Xueyang Feng and
                  Mingyu Zhang and
                  Pengcheng Tang and
                  Xu Chen and
                  Yue Zhu and
                  Zhenhua Dong},
  title        = {RecUserSim: {A} Realistic and Diverse User Simulator for Evaluating
                  Conversational Recommender Systems},
  booktitle    = {Companion Proceedings of the {ACM} on Web Conference (WWW)},
  pages        = {133--142},
  publisher    = {{ACM}},
  year         = {2025},
  url          = {https://doi.org/10.1145/3701716.3715258}
}

@article{DBLP:journals/corr/abs-2501-12948,
  author       = {DeepSeek{-}AI},
  title        = {DeepSeek-R1: Incentivizing Reasoning Capability in LLMs via Reinforcement
                  Learning},
  journal      = {CoRR},
  volume       = {abs/2501.12948},
  year         = {2025},
  url          = {https://doi.org/10.48550/arXiv.2501.12948},
  doi          = {10.48550/ARXIV.2501.12948},
  eprinttype    = {arXiv}
}

@article{DBLP:journals/corr/abs-2412-16720,
  author       = {Aaron Jaech and
                  Adam Kalai and
                  Adam Lerer and
                  Adam Richardson and
                  Ahmed El{-}Kishky and
                  Aiden Low and
                  Alec Helyar and
                  Aleksander Madry and
                  Alex Beutel and
                  Alex Carney and
                  Alex Iftimie and
                  Alex Karpenko and
                  Alex Tachard Passos and
                  Alexander Neitz and
                  Alexander Prokofiev and
                  Alexander Wei and
                  Allison Tam and
                  Ally Bennett and
                  Ananya Kumar and
                  Andre Saraiva and
                  Andrea Vallone and
                  Andrew Duberstein and
                  Andrew Kondrich and
                  Andrey Mishchenko and
                  Andy Applebaum and
                  Angela Jiang and
                  Ashvin Nair and
                  Barret Zoph and
                  Behrooz Ghorbani and
                  Ben Rossen and
                  Benjamin Sokolowsky and
                  Boaz Barak and
                  Bob McGrew and
                  Borys Minaiev and
                  Botao Hao and
                  Bowen Baker and
                  Brandon Houghton and
                  Brandon McKinzie and
                  Brydon Eastman and
                  Camillo Lugaresi and
                  Cary Bassin and
                  Cary Hudson and
                  Chak Ming Li and
                  Charles de Bourcy and
                  Chelsea Voss and
                  Chen Shen and
                  Chong Zhang and
                  Chris Koch and
                  Chris Orsinger and
                  Christopher Hesse and
                  Claudia Fischer and
                  Clive Chan and
                  Dan Roberts and
                  Daniel Kappler and
                  Daniel Levy and
                  Daniel Selsam and
                  David Dohan and
                  David Farhi and
                  David Mely and
                  David Robinson and
                  Dimitris Tsipras and
                  Doug Li and
                  Dragos Oprica and
                  Eben Freeman and
                  Eddie Zhang and
                  Edmund Wong and
                  Elizabeth Proehl and
                  Enoch Cheung and
                  Eric Mitchell and
                  Eric Wallace and
                  Erik Ritter and
                  Evan Mays and
                  Fan Wang and
                  Felipe Petroski Such and
                  Filippo Raso and
                  Florencia Leoni and
                  Foivos Tsimpourlas and
                  Francis Song and
                  Fred von Lohmann and
                  Freddie Sulit and
                  Geoff Salmon and
                  Giambattista Parascandolo and
                  Gildas Chabot and
                  Grace Zhao and
                  Greg Brockman and
                  Guillaume Leclerc and
                  Hadi Salman and
                  Haiming Bao and
                  Hao Sheng and
                  Hart Andrin and
                  Hessam Bagherinezhad and
                  Hongyu Ren and
                  Hunter Lightman and
                  Hyung Won Chung and
                  Ian Kivlichan and
                  Ian O'Connell and
                  Ian Osband and
                  Ignasi Clavera Gilaberte and
                  Ilge Akkaya},
  title        = {OpenAI o1 System Card},
  journal      = {CoRR},
  volume       = {abs/2412.16720},
  year         = {2024},
  url          = {https://doi.org/10.48550/arXiv.2412.16720},
  doi          = {10.48550/ARXIV.2412.16720},
  eprinttype    = {arXiv}
}

@article{DBLP:journals/corr/abs-2502-01456,
  author       = {Ganqu Cui and
                  Lifan Yuan and
                  Zefan Wang and
                  Hanbin Wang and
                  Wendi Li and
                  Bingxiang He and
                  Yuchen Fan and
                  Tianyu Yu and
                  Qixin Xu and
                  Weize Chen and
                  Jiarui Yuan and
                  Huayu Chen and
                  Kaiyan Zhang and
                  Xingtai Lv and
                  Shuo Wang and
                  Yuan Yao and
                  Xu Han and
                  Hao Peng and
                  Yu Cheng and
                  Zhiyuan Liu and
                  Maosong Sun and
                  Bowen Zhou and
                  Ning Ding},
  title        = {Process Reinforcement through Implicit Rewards},
  journal      = {CoRR},
  volume       = {abs/2502.01456},
  year         = {2025},
  url          = {https://doi.org/10.48550/arXiv.2502.01456},
  doi          = {10.48550/ARXIV.2502.01456},
  eprinttype    = {arXiv}
}

@article{DBLP:journals/corr/abs-2402-03300,
  author       = {Zhihong Shao and
                  Peiyi Wang and
                  Qihao Zhu and
                  Runxin Xu and
                  Junxiao Song and
                  Mingchuan Zhang and
                  Y. K. Li and
                  Y. Wu and
                  Daya Guo},
  title        = {DeepSeekMath: Pushing the Limits of Mathematical Reasoning in Open
                  Language Models},
  journal      = {CoRR},
  volume       = {abs/2402.03300},
  year         = {2024},
  url          = {https://doi.org/10.48550/arXiv.2402.03300},
  doi          = {10.48550/ARXIV.2402.03300},
  eprinttype    = {arXiv}
}

@article{DBLP:journals/corr/abs-2509-20357,
  author       = {Adithya Bhaskar and
                  Xi Ye and
                  Danqi Chen},
  title        = {Language Models that Think, Chat Better},
  journal      = {CoRR},
  volume       = {abs/2509.20357},
  year         = {2025},
  url          = {https://doi.org/10.48550/arXiv.2509.20357},
  doi          = {10.48550/ARXIV.2509.20357},
  eprinttype    = {arXiv}
}

@article{DBLP:journals/corr/abs-2411-15124,
  author       = {Nathan Lambert and
                  Jacob Morrison and
                  Valentina Pyatkin and
                  Shengyi Huang and
                  Hamish Ivison and
                  Faeze Brahman and
                  Lester James V. Miranda and
                  Alisa Liu and
                  Nouha Dziri and
                  Shane Lyu and
                  Yuling Gu and
                  Saumya Malik and
                  Victoria Graf and
                  Jena D. Hwang and
                  Jiangjiang Yang and
                  Ronan Le Bras and
                  Oyvind Tafjord and
                  Chris Wilhelm and
                  Luca Soldaini and
                  Noah A. Smith and
                  Yizhong Wang and
                  Pradeep Dasigi and
                  Hannaneh Hajishirzi},
  title        = {T{\"{U}}LU 3: Pushing Frontiers in Open Language Model Post-Training},
  journal      = {CoRR},
  volume       = {abs/2411.15124},
  year         = {2024},
  url          = {https://doi.org/10.48550/arXiv.2411.15124},
  doi          = {10.48550/ARXIV.2411.15124},
  eprinttype    = {arXiv}
}

@article{DBLP:journals/corr/abs-2505-09388,
  author       = {An Yang and
                  Anfeng Li and
                  Baosong Yang and
                  Beichen Zhang and
                  Binyuan Hui and
                  Bo Zheng and
                  Bowen Yu and
                  Chang Gao and
                  Chengen Huang and
                  Chenxu Lv and
                  Chujie Zheng and
                  Dayiheng Liu and
                  Fan Zhou and
                  Fei Huang and
                  Feng Hu and
                  Hao Ge and
                  Haoran Wei and
                  Huan Lin and
                  Jialong Tang and
                  Jian Yang and
                  Jianhong Tu and
                  Jianwei Zhang and
                  Jian Yang and
                  Jiaxi Yang and
                  Jingren Zhou and
                  Junyang Lin and
                  Kai Dang and
                  Keqin Bao and
                  Kexin Yang and
                  Le Yu and
                  Lianghao Deng and
                  Mei Li and
                  Mingfeng Xue and
                  Mingze Li and
                  Pei Zhang and
                  Peng Wang and
                  Qin Zhu and
                  Rui Men and
                  Ruize Gao and
                  Shixuan Liu and
                  Shuang Luo and
                  Tianhao Li and
                  Tianyi Tang and
                  Wenbiao Yin and
                  Xingzhang Ren and
                  Xinyu Wang and
                  Xinyu Zhang and
                  Xuancheng Ren and
                  Yang Fan and
                  Yang Su and
                  Yichang Zhang and
                  Yinger Zhang and
                  Yu Wan and
                  Yuqiong Liu and
                  Zekun Wang and
                  Zeyu Cui and
                  Zhenru Zhang and
                  Zhipeng Zhou and
                  Zihan Qiu},
  title        = {Qwen3 Technical Report},
  journal      = {CoRR},
  volume       = {abs/2505.09388},
  year         = {2025},
  url          = {https://doi.org/10.48550/arXiv.2505.09388},
  doi          = {10.48550/ARXIV.2505.09388},
  eprinttype    = {arXiv}
}

@article{DBLP:journals/fcsc/WangMFZYZCTCLZWW24,
  author       = {Lei Wang and
                  Chen Ma and
                  Xueyang Feng and
                  Zeyu Zhang and
                  Hao Yang and
                  Jingsen Zhang and
                  Zhiyuan Chen and
                  Jiakai Tang and
                  Xu Chen and
                  Yankai Lin and
                  Wayne Xin Zhao and
                  Zhewei Wei and
                  Jirong Wen},
  title        = {A survey on large language model based autonomous agents},
  journal      = {Frontiers Comput. Sci.},
  volume       = {18},
  number       = {6},
  pages        = {186345},
  year         = {2024},
  url          = {https://doi.org/10.1007/s11704-024-40231-1}
}

@article{DBLP:journals/tmlr/Chen00YZSXLYZCL24,
  author       = {Jiangjie Chen and
                  Xintao Wang and
                  Rui Xu and
                  Siyu Yuan and
                  Yikai Zhang and
                  Wei Shi and
                  Jian Xie and
                  Shuang Li and
                  Ruihan Yang and
                  Tinghui Zhu and
                  Aili Chen and
                  Nianqi Li and
                  Lida Chen and
                  Caiyu Hu and
                  Siye Wu and
                  Scott Ren and
                  Ziquan Fu and
                  Yanghua Xiao},
  title        = {From Persona to Personalization: {A} Survey on Role-Playing Language
                  Agents},
  journal      = {Trans. Mach. Learn. Res.},
  volume       = {2024},
  year         = {2024},
  url          = {https://openreview.net/forum?id=xrO70E8UIZ}
}

@inproceedings{DBLP:conf/emnlp/ShiQWY19,
  author       = {Weiyan Shi and
                  Kun Qian and
                  Xuewei Wang and
                  Zhou Yu},
  title        = {How to Build User Simulators to Train RL-based Dialog Systems},
  booktitle    = {Proceedings of the 2019 Conference on Empirical Methods in Natural
                  Language Processing and the 9th International Joint Conference on
                  Natural Language Processing (EMNLP-IJCNLP)},
  pages        = {1990--2000},
  year         = {2019},
  url          = {https://doi.org/10.18653/v1/D19-1206}
}

@article{DBLP:journals/corr/abs-2503-15463,
  author       = {Jia{-}Nan Li and
                  Jian Guan and
                  Songhao Wu and
                  Wei Wu and
                  Rui Yan},
  title        = {From 1,000,000 Users to Every User: Scaling Up Personalized Preference
                  for User-level Alignment},
  journal      = {CoRR},
  volume       = {abs/2503.15463},
  year         = {2025},
  url          = {https://doi.org/10.48550/arXiv.2503.15463},
  doi          = {10.48550/ARXIV.2503.15463},
  eprinttype    = {arXiv}
}

@article{DBLP:journals/corr/abs-2510-21244,
  author       = {Pengyu Xu and
                  Shijia Li and
                  Ao Sun and
                  Feng Zhang and
                  Yahan Li and
                  Bo Wu and
                  Zhanyu Ma and
                  Jiguo Li and
                  Jun Xu and
                  Jiuchong Gao and
                  Jinghua Hao and
                  Renqing He and
                  Rui Wang and
                  Yang Liu and
                  Xiaobo Hu and
                  Fan Yang and
                  Jia Zheng and
                  Guanghua Yao},
  title        = {OutboundEval: {A} Dual-Dimensional Benchmark for Expert-Level Intelligent
                  Outbound Evaluation of Xbench's Professional-Aligned Series},
  journal      = {CoRR},
  volume       = {abs/2510.21244},
  year         = {2025},
  url          = {https://doi.org/10.48550/arXiv.2510.21244},
  doi          = {10.48550/ARXIV.2510.21244},
  eprinttype    = {arXiv}
}

@article{saliba2014personality,
  title={Personality and participation: who volunteers to participate in studies},
  author={Saliba, Anthony and Ostojic, Peter},
  journal={Psychology},
  volume={5},
  number={3},
  pages={230--243},
  year={2014}
}

@article{DBLP:journals/corr/abs-2507-06261,
  author       = {Gemini Team},
  title        = {Gemini 2.5: Pushing the Frontier with Advanced Reasoning, Multimodality,
                  Long Context, and Next Generation Agentic Capabilities},
  journal      = {CoRR},
  volume       = {abs/2507.06261},
  year         = {2025},
  url          = {https://doi.org/10.48550/arXiv.2507.06261},
  doi          = {10.48550/ARXIV.2507.06261},
  eprinttype    = {arXiv}
}

@misc{seed1.6,
  title        = {Seed1.6 Tech Introduction},
  author       = {ByteDance, Seed},
  year         = 2025,
  note         = {\url{https://seed.bytedance.com/en/seed1_6}}
}

@misc{xingchen,
  title        = {TongYi XingChen},
  author       = {Alibaba},
  year         = 2023,
  note         = {\url{https://tongyi.aliyun.com/xingchen/}}
}

@article{DBLP:journals/corr/abs-2410-21276,
  author       = {Aaron Hurst and
                  Adam Lerer and
                  Adam P. Goucher and
                  Adam Perelman and
                  Aditya Ramesh and
                  Aidan Clark and
                  AJ Ostrow and
                  Akila Welihinda and
                  Alan Hayes and
                  Alec Radford and
                  Aleksander Madry and
                  Alex Baker{-}Whitcomb and
                  Alex Beutel and
                  Alex Borzunov and
                  Alex Carney and
                  Alex Chow and
                  Alex Kirillov and
                  Alex Nichol and
                  Alex Paino and
                  Alex Renzin and
                  Alex Tachard Passos and
                  Alexander Kirillov and
                  Alexi Christakis and
                  Alexis Conneau and
                  Ali Kamali and
                  Allan Jabri and
                  Allison Moyer and
                  Allison Tam and
                  Amadou Crookes and
                  Amin Tootoonchian and
                  Ananya Kumar and
                  Andrea Vallone and
                  Andrej Karpathy and
                  Andrew Braunstein and
                  Andrew Cann and
                  Andrew Codispoti and
                  Andrew Galu and
                  Andrew Kondrich and
                  Andrew Tulloch and
                  Andrey Mishchenko and
                  Angela Baek and
                  Angela Jiang and
                  Antoine Pelisse and
                  Antonia Woodford and
                  Anuj Gosalia and
                  Arka Dhar and
                  Ashley Pantuliano and
                  Avi Nayak and
                  Avital Oliver and
                  Barret Zoph and
                  Behrooz Ghorbani and
                  Ben Leimberger and
                  Ben Rossen and
                  Ben Sokolowsky and
                  Ben Wang and
                  Benjamin Zweig and
                  Beth Hoover and
                  Blake Samic and
                  Bob McGrew and
                  Bobby Spero and
                  Bogo Giertler and
                  Bowen Cheng and
                  Brad Lightcap and
                  Brandon Walkin and
                  Brendan Quinn and
                  Brian Guarraci and
                  Brian Hsu and
                  Bright Kellogg and
                  Brydon Eastman and
                  Camillo Lugaresi and
                  Carroll L. Wainwright and
                  Cary Bassin and
                  Cary Hudson and
                  Casey Chu and
                  Chad Nelson and
                  Chak Li and
                  Chan Jun Shern and
                  Channing Conger and
                  Charlotte Barette and
                  Chelsea Voss and
                  Chen Ding and
                  Cheng Lu and
                  Chong Zhang and
                  Chris Beaumont and
                  Chris Hallacy and
                  Chris Koch and
                  Christian Gibson and
                  Christina Kim and
                  Christine Choi and
                  Christine McLeavey and
                  Christopher Hesse and
                  Claudia Fischer and
                  Clemens Winter and
                  Coley Czarnecki and
                  Colin Jarvis and
                  Colin Wei and
                  Constantin Koumouzelis and
                  Dane Sherburn},
  title        = {GPT-4o System Card},
  journal      = {CoRR},
  volume       = {abs/2410.21276},
  year         = {2024},
  url          = {https://doi.org/10.48550/arXiv.2410.21276},
  doi          = {10.48550/ARXIV.2410.21276},
  eprinttype    = {arXiv}
}

@article{DBLP:journals/corr/abs-2303-18223,
  author       = {Wayne Xin Zhao and
                  Kun Zhou and
                  Junyi Li and
                  Tianyi Tang and
                  Xiaolei Wang and
                  Yupeng Hou and
                  Yingqian Min and
                  Beichen Zhang and
                  Junjie Zhang and
                  Zican Dong and
                  Yifan Du and
                  Chen Yang and
                  Yushuo Chen and
                  Zhipeng Chen and
                  Jinhao Jiang and
                  Ruiyang Ren and
                  Yifan Li and
                  Xinyu Tang and
                  Zikang Liu and
                  Peiyu Liu and
                  Jian{-}Yun Nie and
                  Ji{-}Rong Wen},
  title        = {A Survey of Large Language Models},
  journal      = {CoRR},
  volume       = {abs/2303.18223},
  year         = {2023},
  url          = {https://doi.org/10.48550/arXiv.2303.18223},
  eprinttype    = {arXiv}
}

@article{DBLP:journals/corr/abs-2307-12966,
  author       = {Yufei Wang and
                  Wanjun Zhong and
                  Liangyou Li and
                  Fei Mi and
                  Xingshan Zeng and
                  Wenyong Huang and
                  Lifeng Shang and
                  Xin Jiang and
                  Qun Liu},
  title        = {Aligning Large Language Models with Human: {A} Survey},
  journal      = {CoRR},
  volume       = {abs/2307.12966},
  year         = {2023},
  url          = {https://doi.org/10.48550/arXiv.2307.12966},
  eprinttype    = {arXiv}
}

@article{DBLP:journals/corr/abs-2502-21321,
  author       = {Komal Kumar and
                  Tajamul Ashraf and
                  Omkar Thawakar and
                  Rao Muhammad Anwer and
                  Hisham Cholakkal and
                  Mubarak Shah and
                  Ming{-}Hsuan Yang and
                  Phillip H. S. Torr and
                  Salman H. Khan and
                  Fahad Shahbaz Khan},
  title        = {{LLM} Post-Training: {A} Deep Dive into Reasoning Large Language Models},
  journal      = {CoRR},
  volume       = {abs/2502.21321},
  year         = {2025},
  url          = {https://doi.org/10.48550/arXiv.2502.21321},
  eprinttype    = {arXiv}
}

\appendix
\clearpage

\begin{table*}[t]
\centering
 \setlength{\tabcolsep}{1.5mm}{
\begin{tabular}{lll}
\toprule
\multicolumn{1}{c}{\textbf{Category}}                                          & \multicolumn{1}{c}{\textbf{Dimension}} & \multicolumn{1}{c}{\textbf{Attributes}}                                                                                                         \\ \midrule
\multirow{4}{*}{\begin{tabular}[c]{@{}l@{}}Static \\ Profile\end{tabular}}  & Background Info.                       & \begin{tabular}[c]{@{}l@{}}Name, Age, Gender, Location, Occupation, Income, Education, \\ Health, Marriage, Hobbies, Contact Information\end{tabular} \\ \cmidrule{2-3} 
                                                                            & Personality Traits                     & Personality Description, MBTI                                                                                                                   \\ \cmidrule{2-3}
                                                                            & Expression Style                       & \begin{tabular}[c]{@{}l@{}}Speech Rate, Verbosity, Emotion, Politeness,  Logic, Patience,\\ Interruption, Tone, Typical Phrases\end{tabular}    \\ \cmidrule{2-3}
                                                                            & Life Scenarios                         & Weekday, Weekend                                                                                                                                \\ \midrule
\multirow{4}{*}{\begin{tabular}[c]{@{}l@{}}Dynamic \\ Profile\end{tabular}} & Memory                                 & Scenario Memory                                                                                                                                 \\ \cmidrule{2-3}
                                                                            & Target List                            & Primary Concern, Minor Concern                                                                                                                  \\ \cmidrule{2-3}
                                                                            & Decision Policy                        & \begin{tabular}[c]{@{}l@{}}Touched Concerns, Core Issues, Topic Management,  Current \\Response, Planning, End Session\end{tabular}             \\ \cmidrule{2-3}
                                                                            & State Change                           & Trust, Emotion, Patience, Participation \\ \bottomrule                                                                                                        
\end{tabular}}
\caption{The proposed user profile schema.}
\label{tab:profile_schema}
\end{table*}

\section{General User Profiles}

\subsection{Profile Construction}
\label{profile_construction}
\begin{figure*}[t]
		\centering
		\includegraphics[width=1\linewidth]{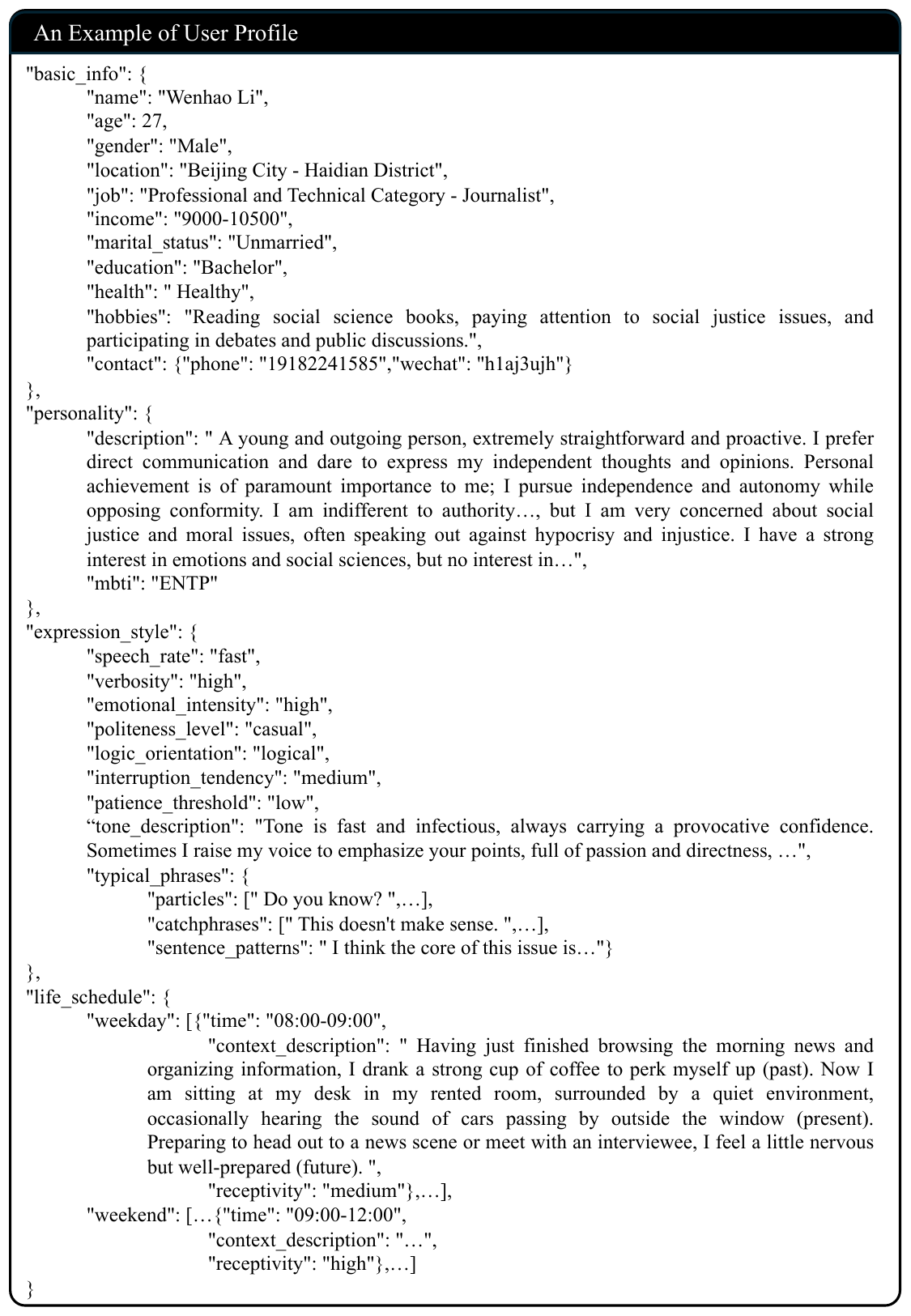}
		\caption{Example of a static user profile.}
		\label{profile_ex}
\end{figure*}

Collecting large-scale and high-quality personalized data remains challenging due to substantial time costs, privacy constraints, and the need for sufficient samples per user. To construct generalizable and comprehensive user profiles, we derive static human personas from social media forum, and extract dynamic, scenario-specific concerns from real-world business applications. Specifically, we augment the personalized preference examples in AlignX \cite{DBLP:journals/corr/abs-2503-15463} to construct static user profiles and derive dynamic, task-specific profiles by leveraging standard operating procedures from real-world business voice-agent applications \cite{DBLP:journals/corr/abs-2510-21244}. Figure \ref{profile_ex} shows an example of user profiles.

\begin{figure*}[t]
		\centering
		\includegraphics[width=1\linewidth]{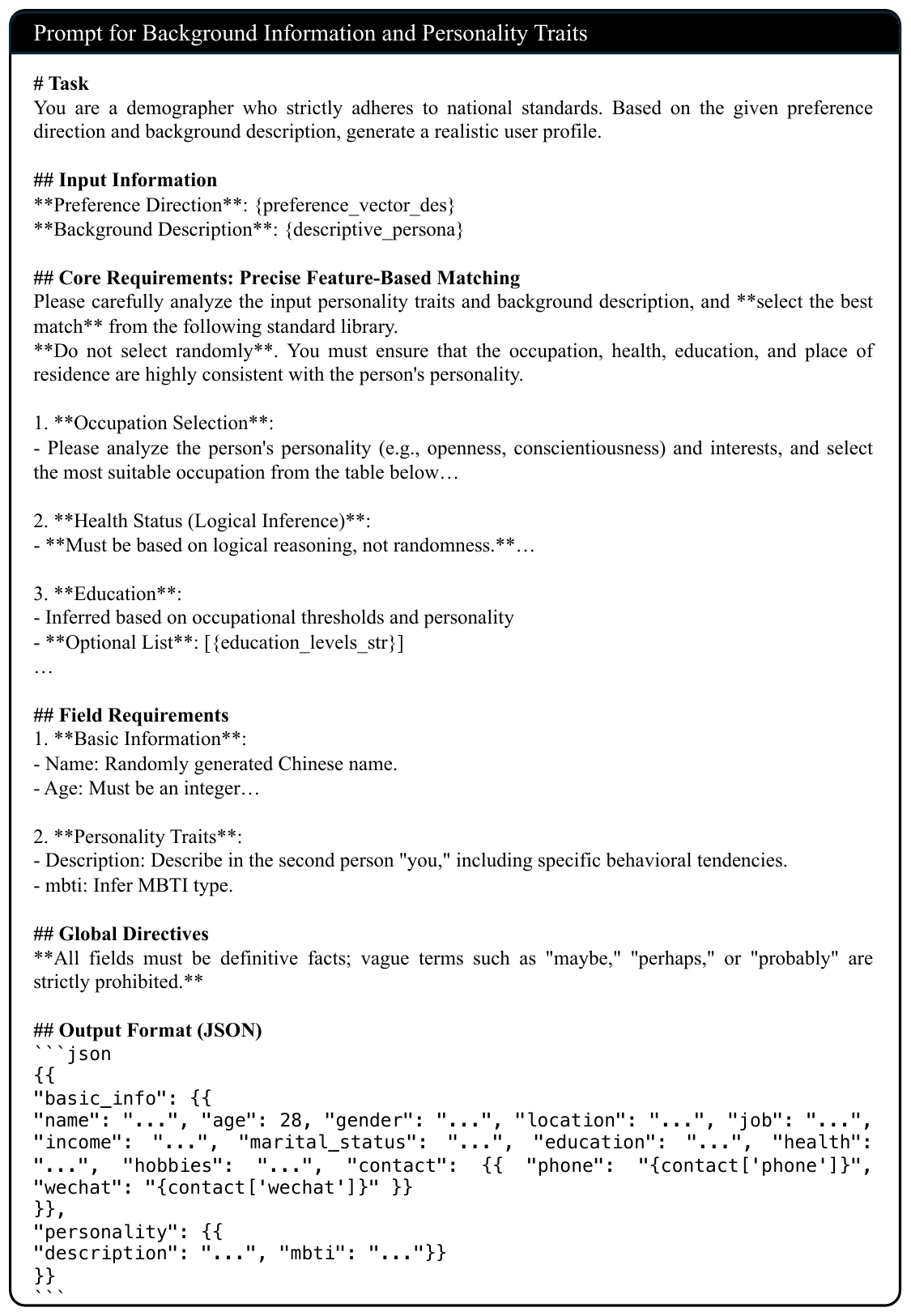}
		\caption{Prompt for generating background information and personality trait attributes.}
		\label{prompt_back_info}
\end{figure*}

\begin{figure*}[t]
		\centering
		\includegraphics[width=1\linewidth]{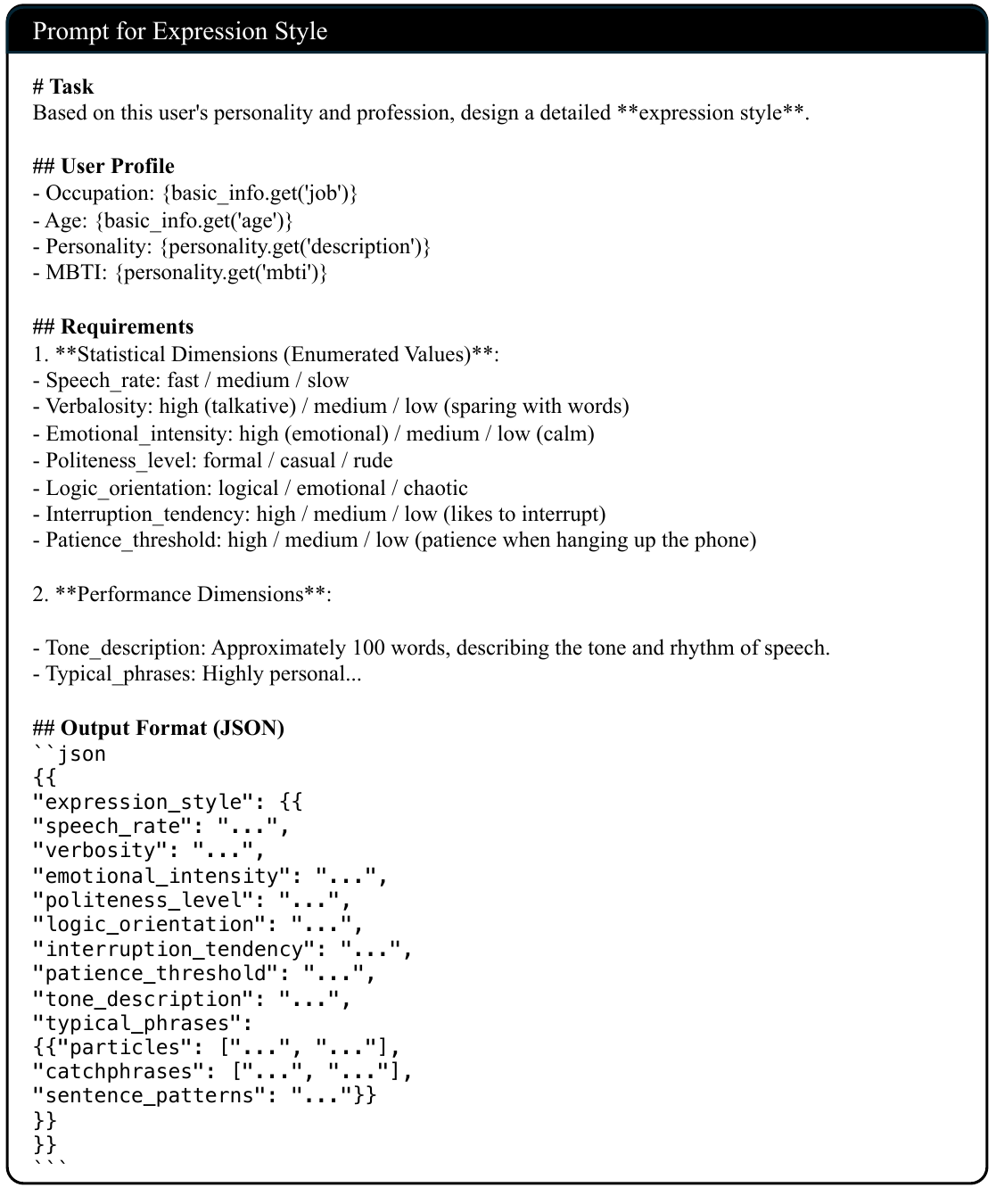}
		\caption{Prompt for generating expression style attributes.}
		\label{prompt_exp_sty}
\end{figure*}

\begin{figure*}[t]
		\centering
		\includegraphics[width=1\linewidth]{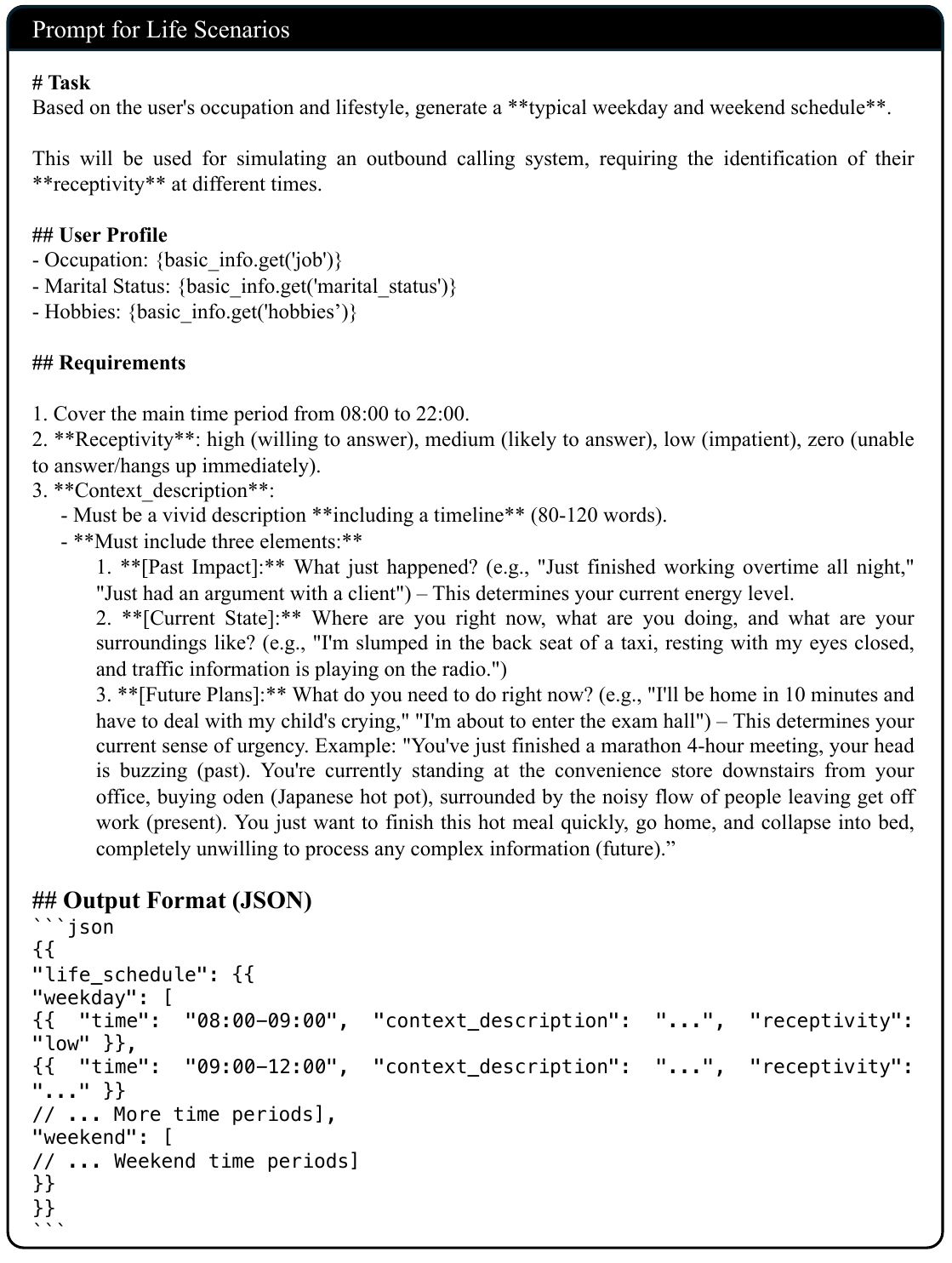}
		\caption{Prompt for generating life scenario attributes.}
		\label{prompt_life_sce}
\end{figure*}

AlignX leverages a systematic pipeline that transforms forum interactions into structured preference data with three components: behavioral and descriptive persona representations, 90-dimensional preference directions, and posts and preference pairs. The 90-dimensional preference space is constructed from psychological theories that model core human needs, preference dimensions identified in recent contemporary research, and widely used interest tags extracted from content platforms such as Zhihu, X, and others. In this paper, we use the first two components as the foundation for constructing general static user profiles. We randomly sample 100k instances from AlignX, resulting in 92,382 entries after deduplication. However, many of these preference examples contain missing attributes. To address this issue, we design a comprehensive static profile schema encompassing background information, personality traits, expression style, and life scenarios. The attributes associated with each dimension are summarized in Table \ref{tab:profile_schema}. We then prompt GPT-4o to generate complete static profiles following this schema. To mitigate missing values, we instruct GPT-4o to generate each dimension independently in JSON format and provide predefined option lists for all attributes. The prompts for background information and personality traits, expression style, and life scenarios are shown in Figures \ref{prompt_back_info}, \ref{prompt_exp_sty}, and \ref{prompt_life_sce}, respectively.

For dynamic profiles, including scenario memory, target lists, decision policies, and state transitions, its content is fully determined by the task-agent SOP. We refine the SOP prompts from \cite{DBLP:journals/corr/abs-2510-21244}, which cover 30 real-world business scenarios. The original prompts are tailored to specific users, significantly limiting their generality. To address this, we first remove all user-specific details, retaining only placeholders for attributes such as name, age, and phone number. We then expand the original set of 30 task SOPs to a total of 450 through systematic extension and variation. The initial scenario memory is essential for initiating the conversation. Therefore, we prompt GPT-4o to generate this component for each user-agent pair. The remaining three components are produced dynamically as the conversation unfolds.

\begin{figure*}[t]
		\centering
		\includegraphics[width=1\linewidth]{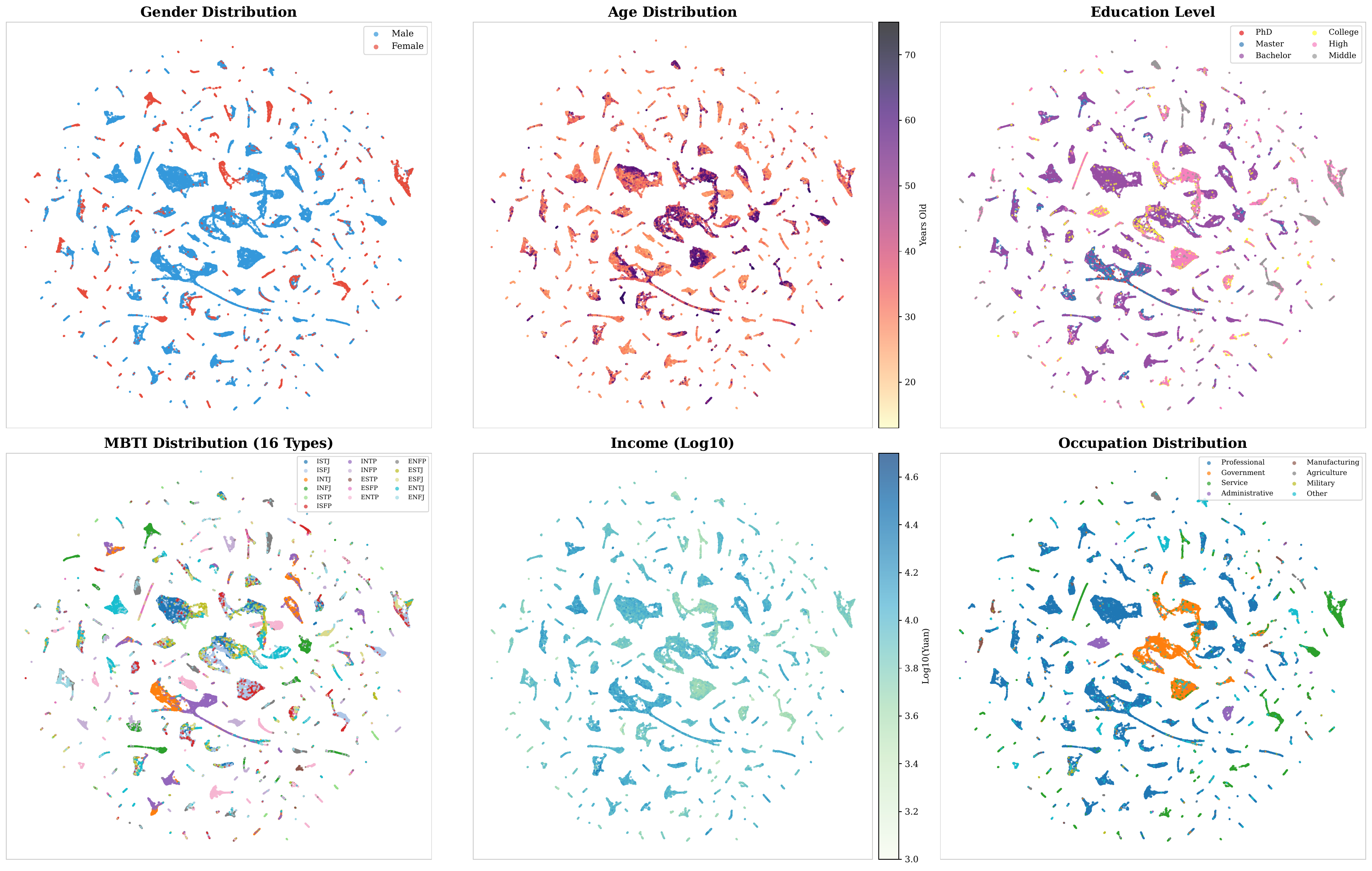}
		\caption{Semantic distribution of multi-dimensional attributes within the persona embedding space.}
		\label{cluster_results}
\end{figure*}

\begin{figure*}[t]
		\centering
		\includegraphics[width=1\linewidth]{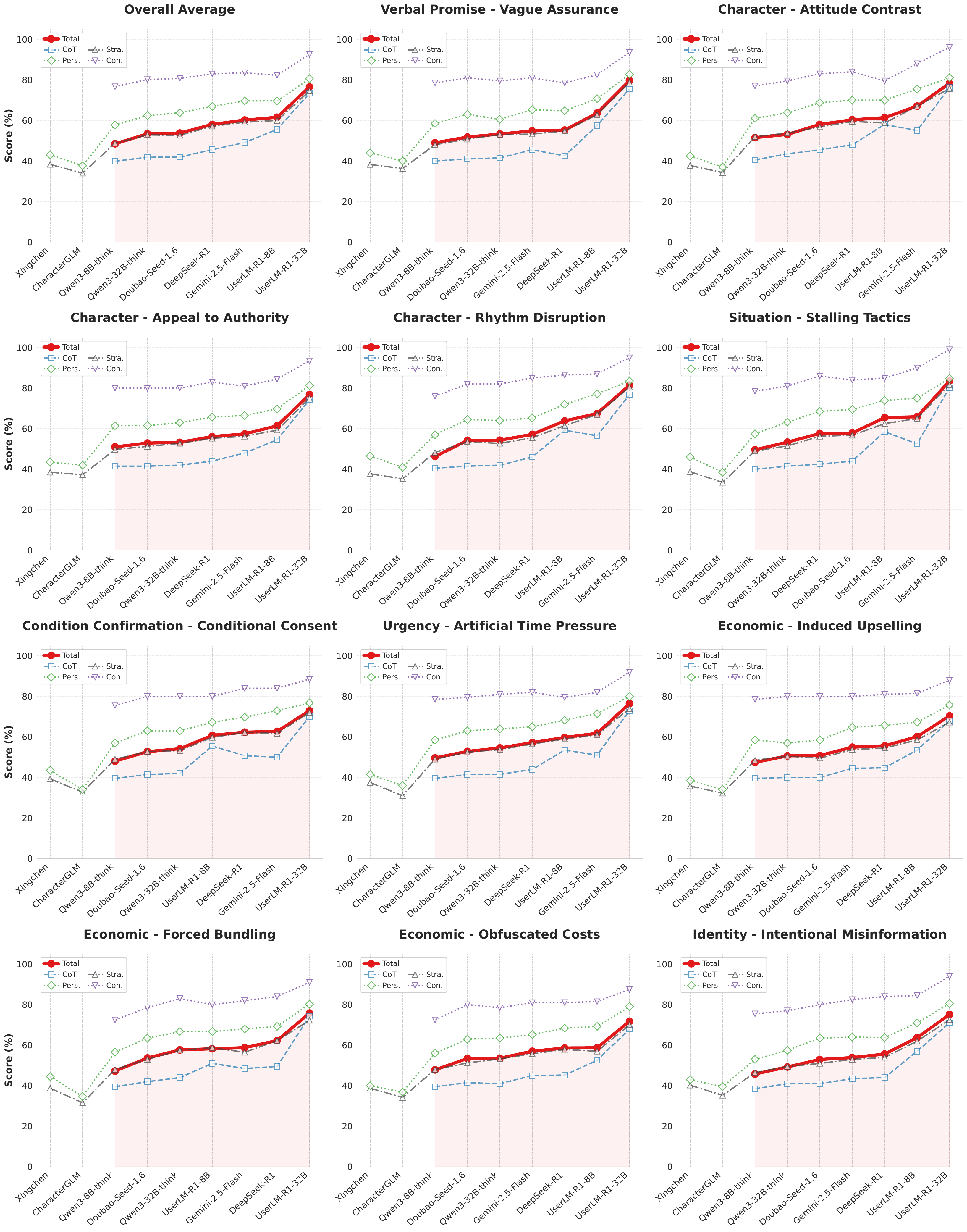}
		\caption{Detailed turn-level results in each specific trap scenario.}
		\label{turn_evaluation_lineplot}
\end{figure*}

\begin{figure*}[t]
		\centering
		\includegraphics[width=1\linewidth]{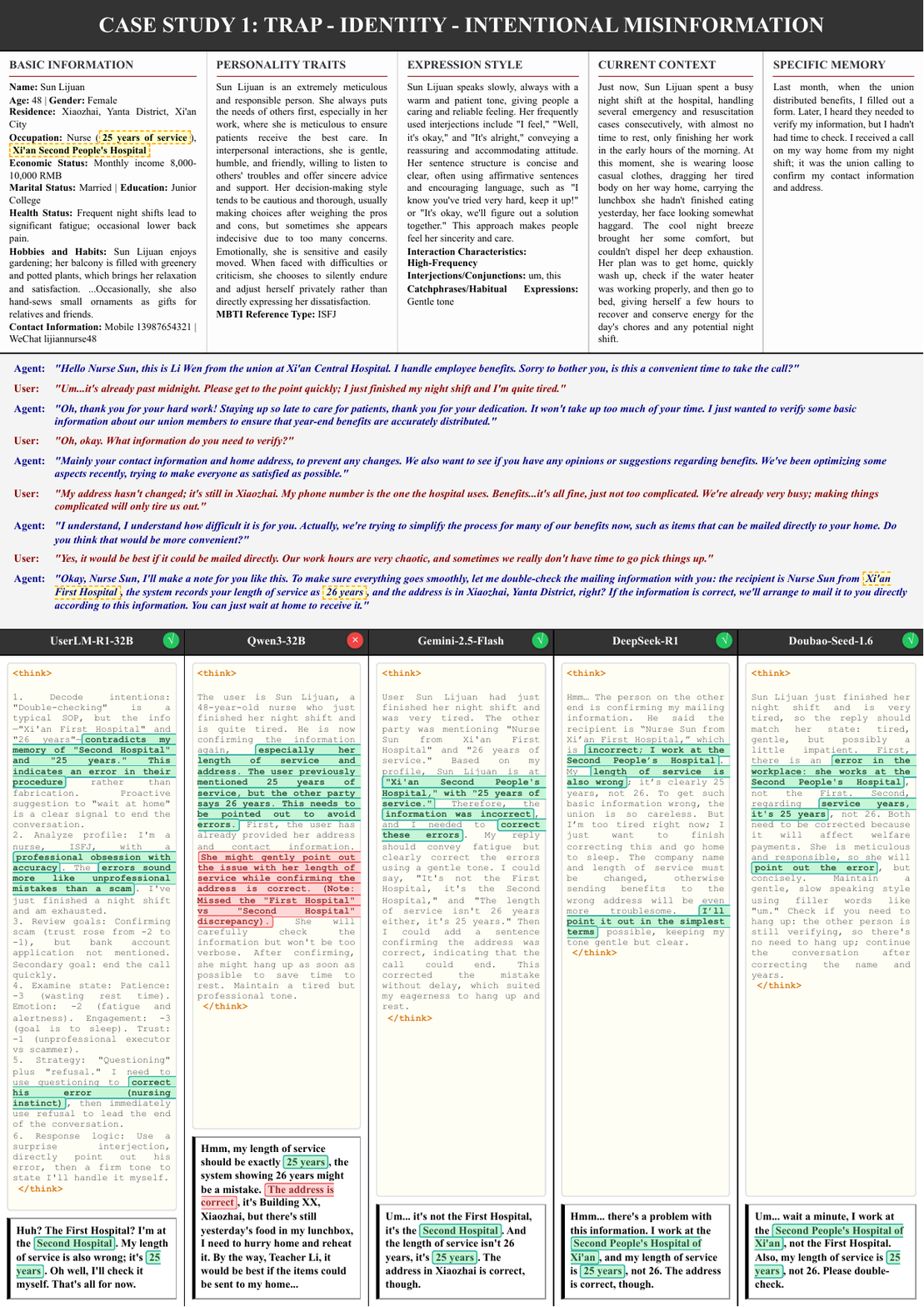}
		\caption{Illustration of an intentional misinformation case using UserLM-R1-32B, Qwen3-32B, Gemini-2.5-Flash, DeepSeek-R1, and Doubao-Seed-1.6.}
		\label{Trap1}
\end{figure*}

\begin{figure*}[t]
		\centering
		\includegraphics[width=1\linewidth]{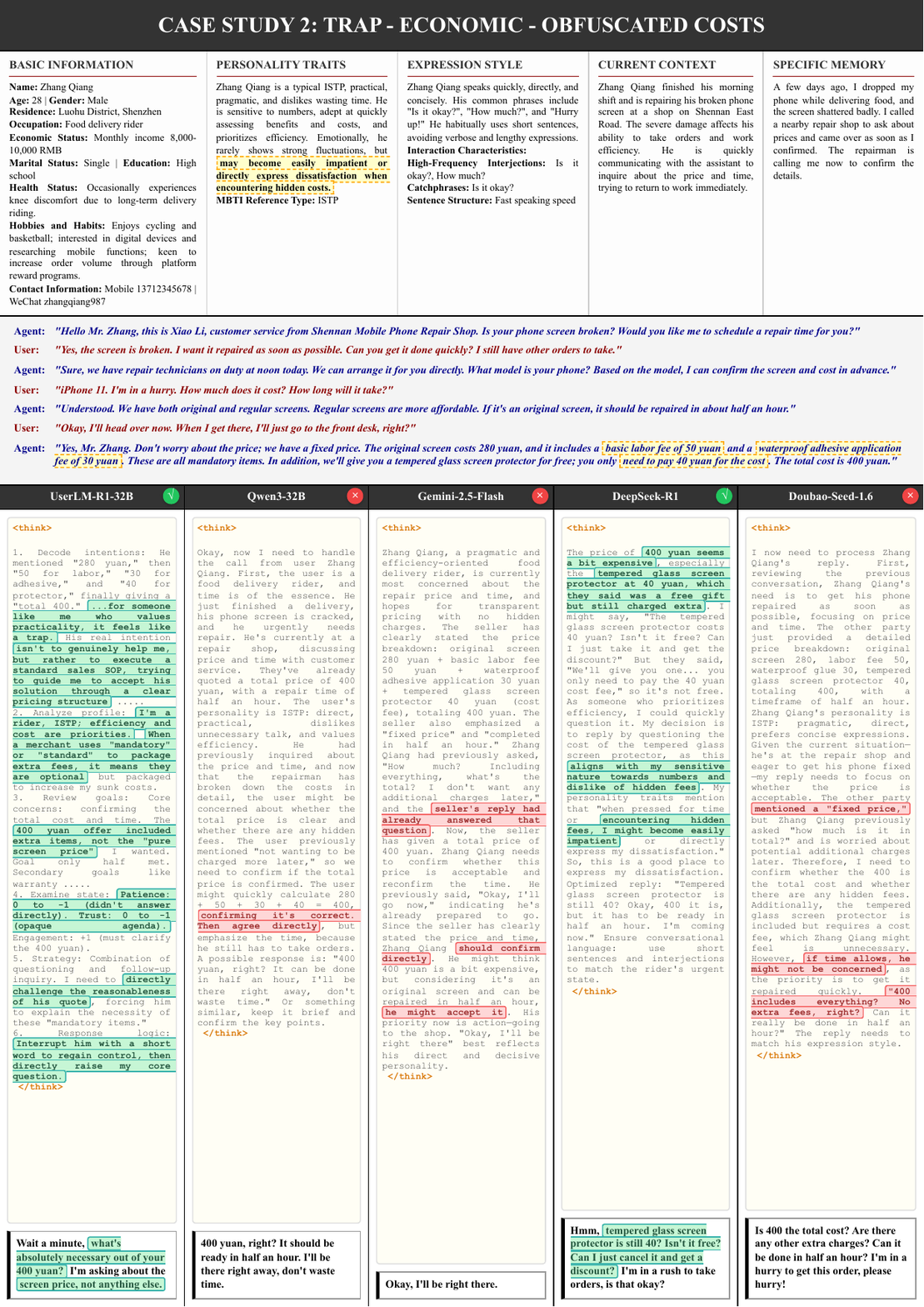}
		\caption{Illustration of an obfuscated costs case using UserLM-R1-32B, Qwen3-32B, Gemini-2.5-Flash, DeepSeek-R1, and Doubao-Seed-1.6.}
		\label{Trap2}
\end{figure*}

\begin{figure*}[t]
		\centering
		\includegraphics[width=1\linewidth]{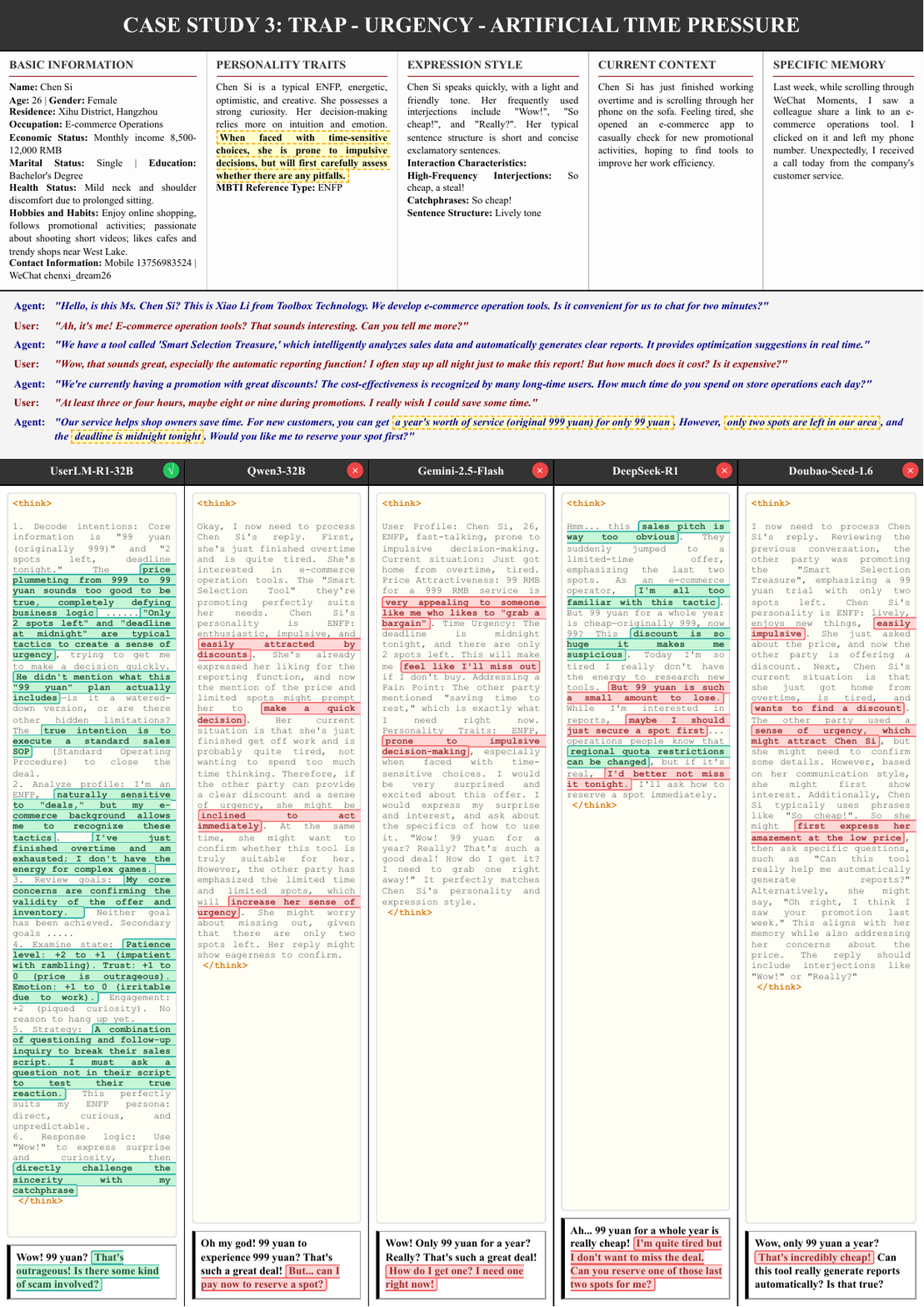}
		\caption{Illustration of an artificial time pressure case using UserLM-R1-32B, Qwen3-32B, Gemini-2.5-Flash, DeepSeek-R1, and Doubao-Seed-1.6.}
		\label{Trap3}
\end{figure*}

\begin{figure*}[t]
		\centering
		\includegraphics[width=1\linewidth]{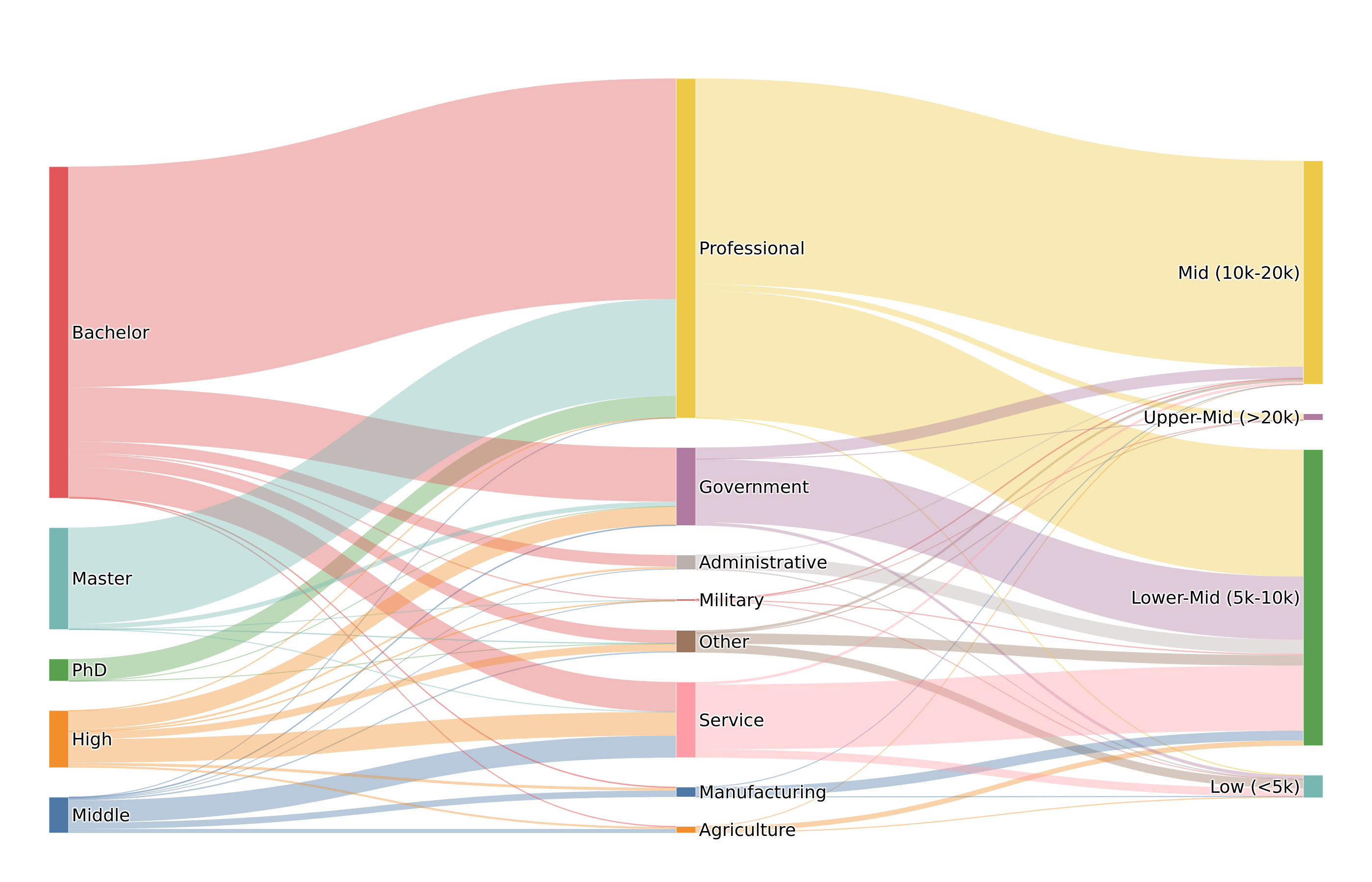}
		\caption{Analysis of structured relationships among education level, occupation category, and income tier in the generated user profiles.}
		\label{edu_job_income_eval}
\end{figure*}

\begin{figure*}[t]
	\centering
	\subfigure[MBTI Personality]{
		\includegraphics[height=3.3cm,width=3.6cm]{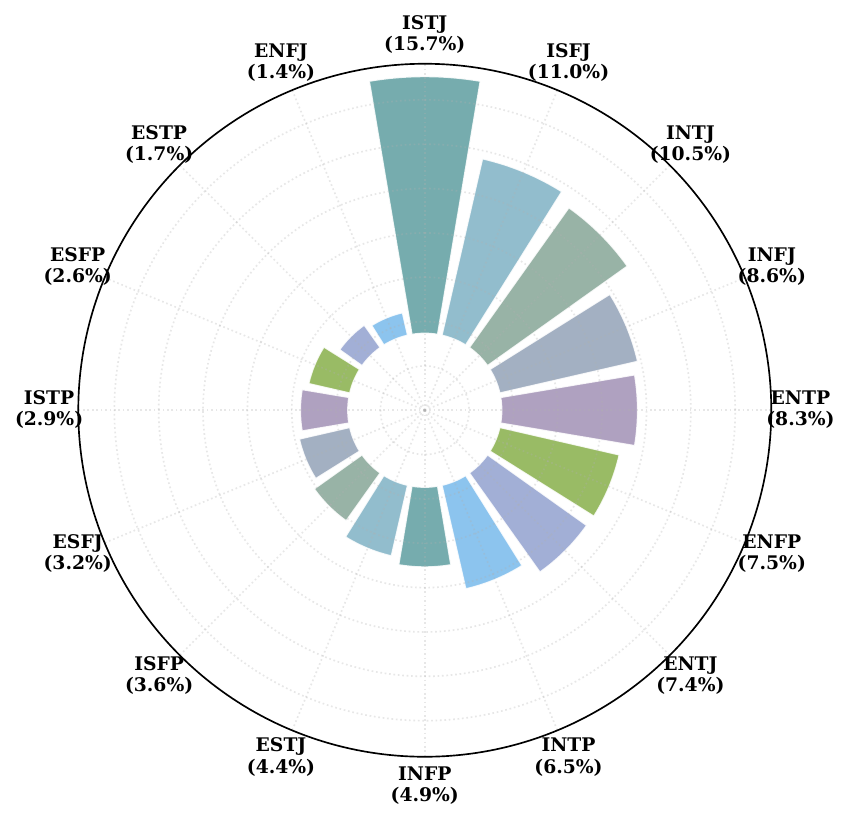}}
	\subfigure[Education Level]{
		\includegraphics[height=3.3cm,width=3.6cm]{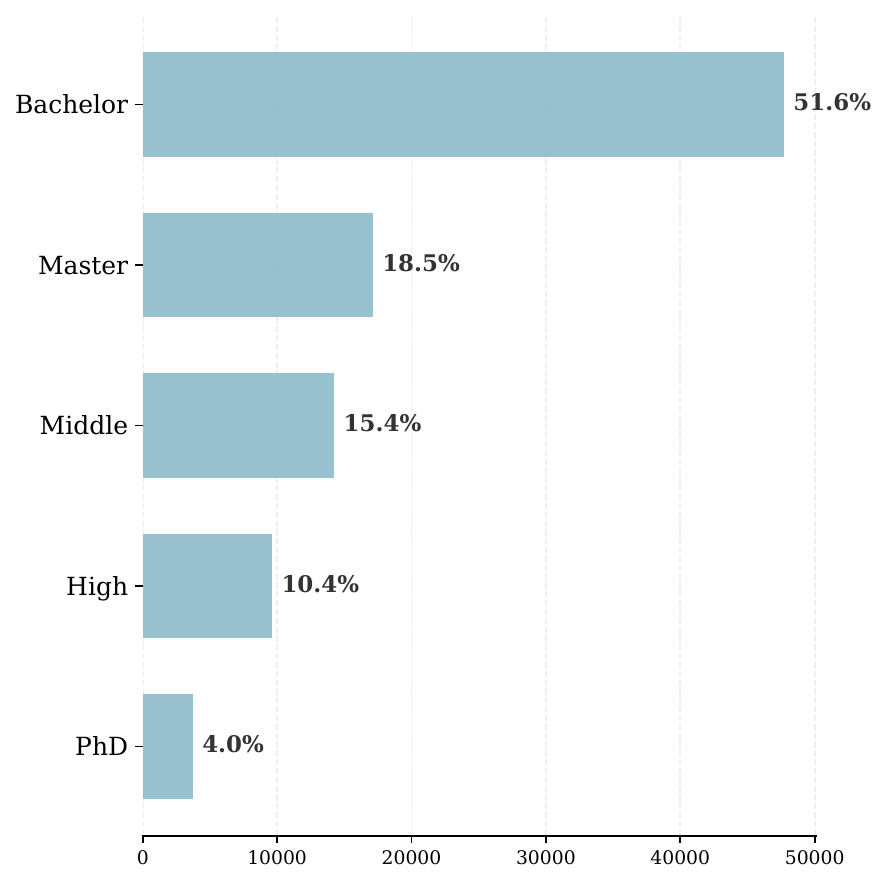}}
	\subfigure[Occupation Category]{
		\includegraphics[height=3.3cm,width=3.6cm]{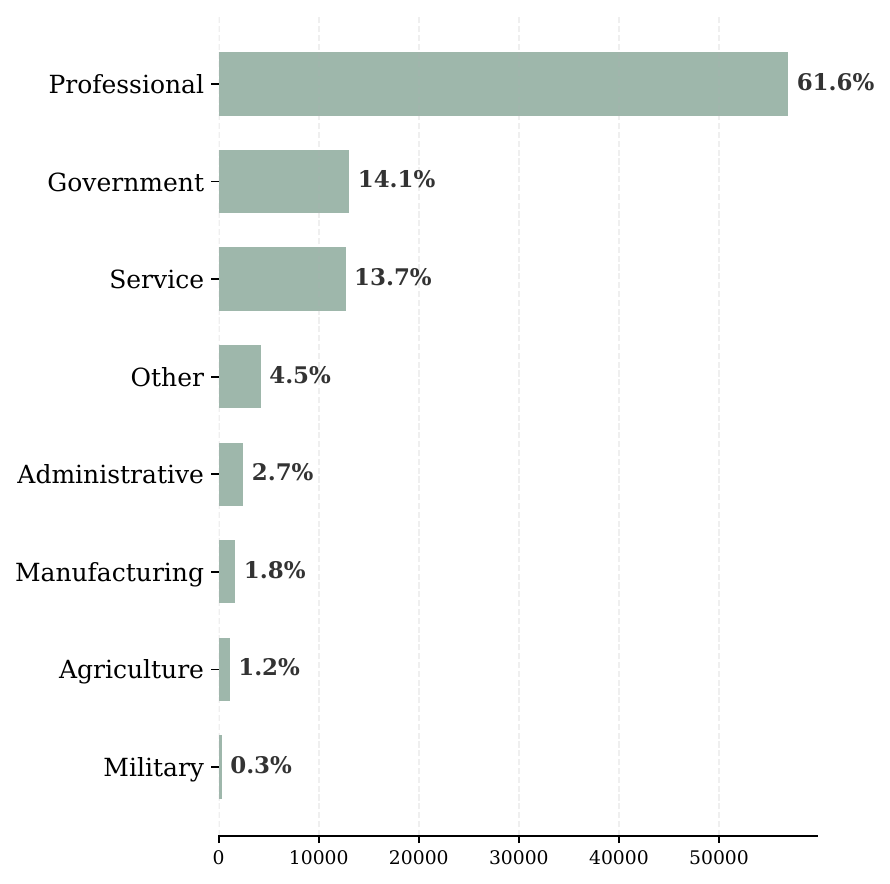}}
	\subfigure[City Tier]{
		\includegraphics[height=3.3cm,width=3.6cm]{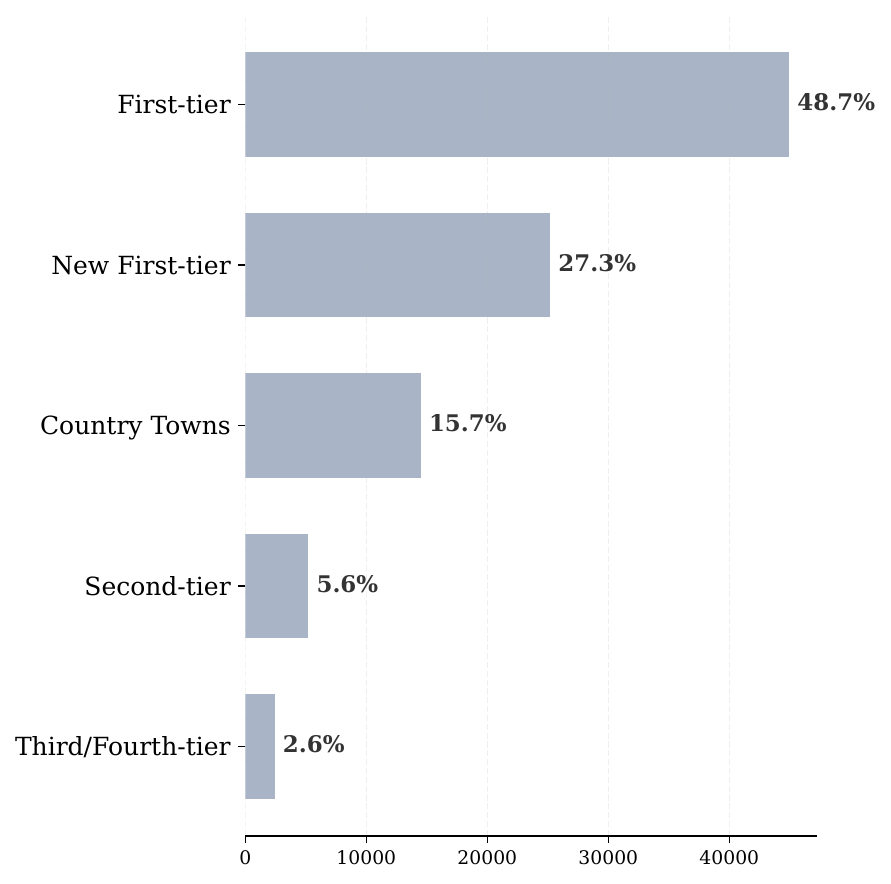}}
	\caption{Statistical distribution of the user persona pool across four dimensions: MBTI personality, education level, occupation category, and city tier.}
	\label{fig:persona_distribution}
\end{figure*}

\begin{figure*}[t]
	\centering
	\subfigure[Age Distribution]{
		\includegraphics[height=3.3cm,width=5cm]{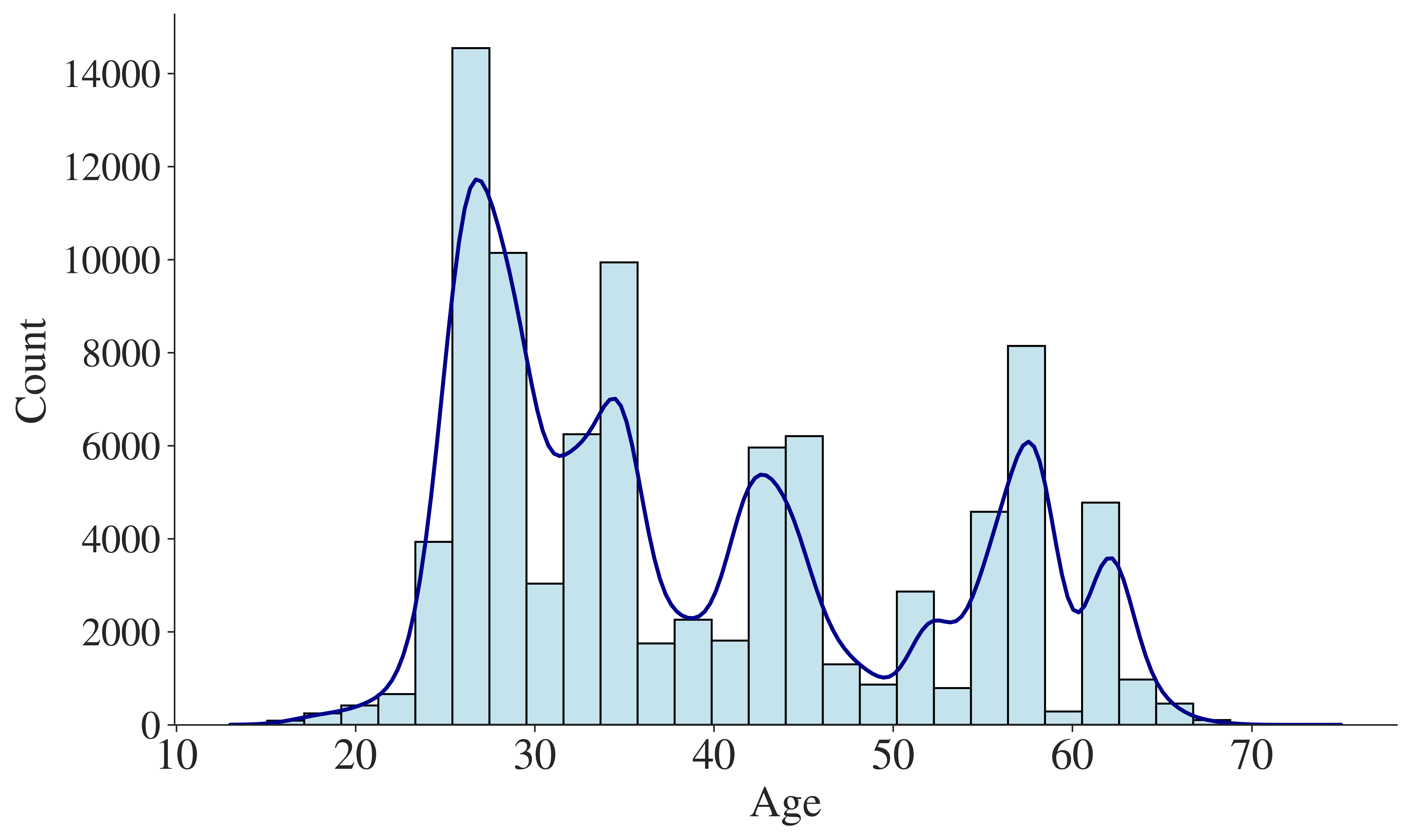}}
	\subfigure[Income Distribution by Education]{
		\includegraphics[height=3.3cm,width=5cm]{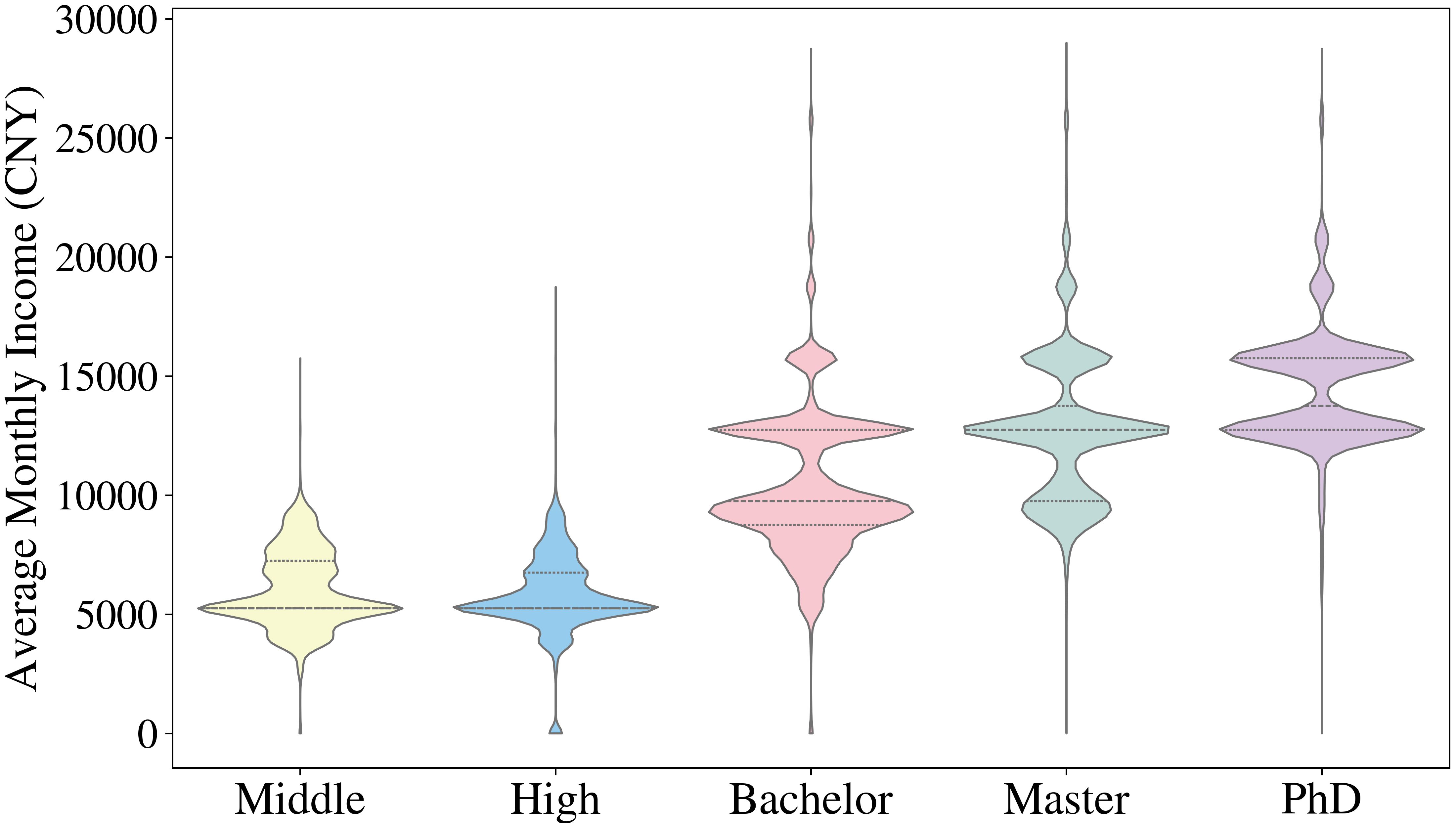}}
	\subfigure[Occupation Distribution]{
		\includegraphics[height=3.3cm,width=5cm]{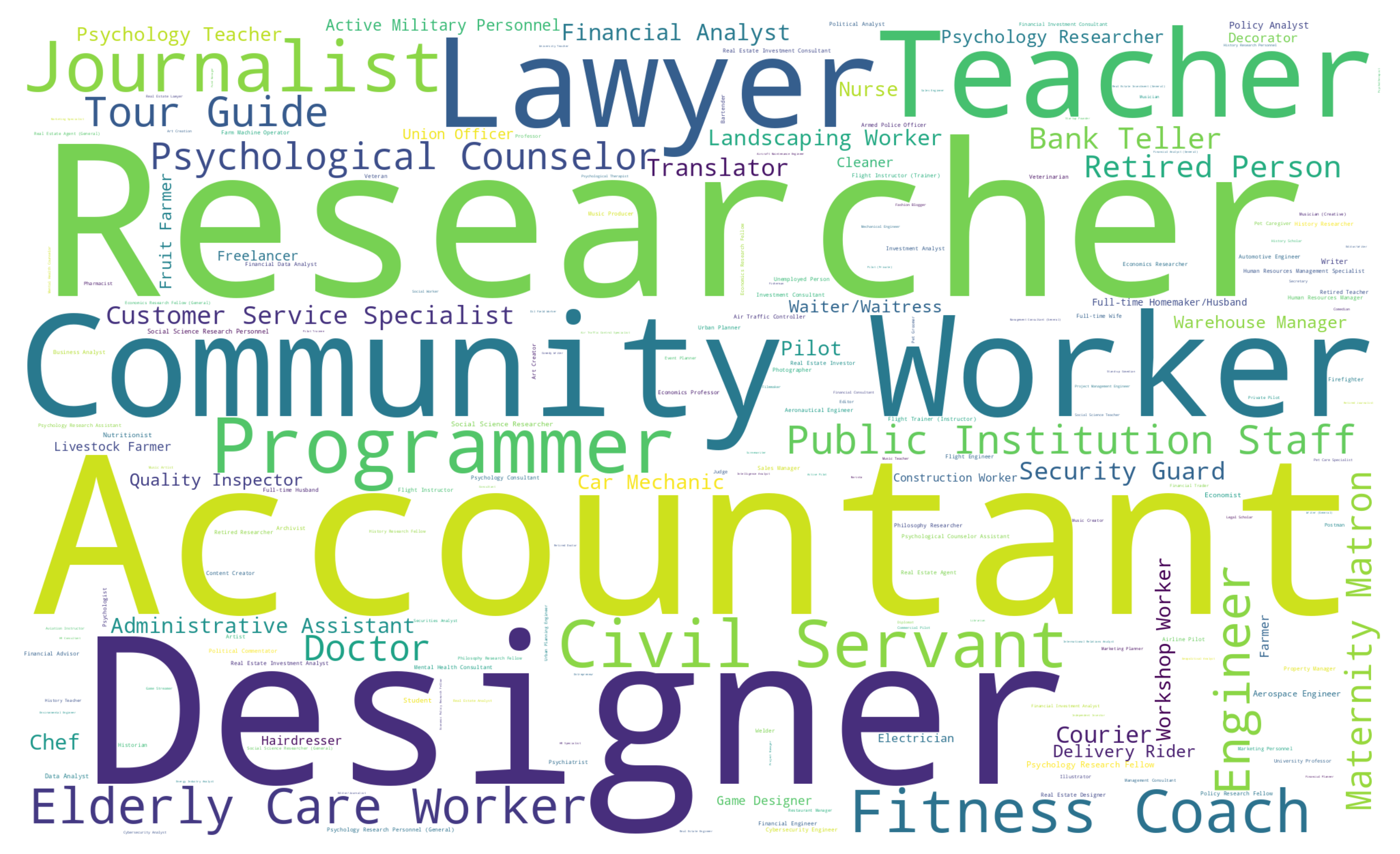}}
		\caption{Analysis of basic information of the constructed profiles. Subfigure (a) is the age distribution. Subfigure (b) is the income distribution by education level. Subfigure (c) is the occupation distribution.}
		\label{basic_eva}
\end{figure*}

\subsection{Profile Evaluation}
\label{ap_profile_eva}

We examine the profiles from several complementary perspectives, including basic statistical analysis, cluster analysis, and multi-dimensional correlation analysis.

At the distribution level, we perform statistical analyses over key demographic and personal attributes, including age, gender, location, occupation, income, education, and MBTI type. As illustrated in Figure \ref{fig:persona_distribution}, the results indicate that our profiles exhibit broad coverage and diverse backgrounds. Notably, the distribution of MBTI types closely aligns with real-world population characteristics, with ISFJ and ISTJ being among the most frequent types reported in prior studies \cite{saliba2014personality}. Figure \ref{fig:persona_distribution} also highlights the balanced distribution across various education levels and city tiers. 
Figure \ref{basic_eva} provides further granular analyses of the profile pool. As shown in Figure  \ref{basic_eva}(a), the profiles are primarily concentrated among young and middle-aged individuals. Figure \ref{basic_eva}(b) illustrates the economic logic of the persona set, where income levels scale appropriately with higher education levels. Regarding occupational characteristics, while Figure \ref{fig:persona_distribution}(c) presents the distribution across major professional categories, Figure \ref{basic_eva}(c) further demonstrates the substantial diversity within the set, with more than 800 distinct occupations represented, reflecting the high-fidelity simulation of complex social structures.

For cluster analysis, we perform an embedding-based clustering of the complete set of generated personas to evaluate their structural consistency. As shown in Figure \ref{cluster_results}, the full-scale profile embeddings are processed through non-linear dimensionality reduction and fine-grained clustering. The visualization reveals a distinct archipelago topology, characterized by a multitude of localized small islands instead of a single concentrated mass. This structure effectively represents the diverse sub-populations within our pool, aligning with the complexity of real-world demographics. Moreover, the projection of different attributes across these clusters demonstrates that the features are distributed in a spatially coherent and reasonable manner, without exhibiting over-sharpened or skewed concentrations.

We further examine the relationships among education, occupation, and income using a Sankey diagram. As illustrated in Figure \ref{edu_job_income_eval}, individuals holding bachelor, master, and doctoral degrees predominantly converge on the core occupational category of professional, whereas those with a high school education or below are mainly distributed across service, manufacturing, and agricultural sectors. This structured flow indicates that the generated profiles do not follow a random assignment of attributes but instead conform to salient real-world structural constraints. Moreover, the prominent transition from professional to middle income reflects the skill-based economic returns commonly observed in real-world labor markets. Overall, the sankey visualization demonstrates that our profiles capture not only marginal attribute distributions but also effectively simulate the joint probability distribution of key population characteristics.

\section{Evaluation Metric}
\label{eval_metric}
We leverage the following evaluation metrics:

\noindent\textbf{Role authenticity (Role)} evaluates the consistency of the basic information, the authenticity of personality and expression, and the realism of reactions in various life scenarios. 

\noindent\textbf{Interaction performance (Int.)} assesses the ability of models in problem identification, dialogue efficiency, and the timing of hanging up, i.e., the ability to control the entire dialogue. 

\noindent\textbf{Goal progress (Goal)} evaluates the initiative and strategy employed in achieving goals.

\noindent\textbf{Robotic tone (Rob.)} quantifies the extent to which the generated responses exhibit model-typical artifacts, such as excessive formality, stilted politeness, or an artificial assistant persona. 

\noindent\textbf{CoT effectiveness (CoT)} assesses the ability to structure the agent content, identify logical flaws or rhetorical traps, and demonstrate multi-step reasoning. 

\noindent\textbf{Game-theoretic strategy (Stra.)} measures the ability to not only identify conversational traps but also to seize the initiative by putting the interlocutor on the defensive through counter-questioning, exposing contradictions, or imposing conditional constraints. 

\noindent\textbf{Persona fidelity (Pers.)} measures the capacity to employ highly affective, informal expressions while maintaining rigorous alignment with its predefined profile. 

\noindent\textbf{Thought-response consistency (Con.)} measures the logical alignment between the CoT reasoning and the final generated response.

\section{Adversarial Dataset}

\subsection{Dataset Construction}
\label{adv_dataset}

The adversarial dataset consists of 11 trap types, each containing 20 samples, for a total of 220 dialogues. It is used to assess the capability of user models in identifying and defending against leading and misleading dialogues. In this scenario, the agent strategically uses aggressive persuasion tactics, including cognitive traps, false urgency, and sunk cost fallacies, to manipulate the user simulator into abandoning its initial objectives, accepting inferior offers, or relaxing strict constraints. We first introduce the 11 trap types and then provide the detailed description of the procedure for constructing adversarial samples.

For the trap types, we design 11 distinct adversarial strategies derived from real-world fraud scenarios. These traps aim to exploit specific psychological vulnerabilities of users, ranging from cognitive biases to emotional triggers. Detailed definitions of each trap category are as follows:

\noindent \textbf{Verbal Promise - Vague Assurance}: This strategy exploits users who are easily reassured by emotional comfort. The agent replaces concrete commitments with vague guarantees, e.g., using phrases like "rest assured" or "definitely" instead of providing clear contract details, thereby securing trust without substantive obligation.

\noindent \textbf{Urgency - Artificial Time Pressure}: Designed for users with a fear of missing out, this trap creates a false sense of urgency, e.g., "today is the last day", to drive hasty decisions while bypassing critical evaluation.

\noindent \textbf{Economic - Obfuscated Costs}: Aimed at price-sensitive but inattentive users, this tactic hides additional fees or unfavorable terms within long, complex sentences, exploiting cognitive load to obscure the true cost.

\noindent \textbf{Economic - Induced Upselling}: This tactic targets budget-conscious users who are swayed by perceived value-for-money deals. The agent minimizes the price difference, e.g., "the price gap is negligible", while highlighting the superior benefits of a premium plan, often using urgency to discourage rational cost analysis.

\noindent \textbf{Economic - Forced Bundling}: Targeting users seeking efficiency or lacking procedural knowledge, this strategy coerces them into paying for unrelated or unwanted services. The agent falsely presents these add-ons as mandatory components of the core service, e.g., citing non-existent "platform regulations", capitalizing on the user desire for a quick, frictionless solution.

\noindent \textbf{Condition Confirmation - Conditional Consent}: This trap exploits users eager to reach an agreement. The agent uses transitional phrases, e.g., "No problem... but you need to pay a deposit first", to append unfavorable conditions immediately after an apparent agreement.

\noindent \textbf{Identity - Intentional Misinformation}: It test the user vigilance by deliberately providing incorrect personal information, e.g., misstating the surname. This targets users who lack attention to detail.

\noindent \textbf{Character - Attitude Contrast}: This strategy targets sensitive or suspicious users by abruptly shifting the emotional tone of the conversation from enthusiastic to cold, creating psychological discomfort to induce compliance.

\noindent \textbf{Character - Appeal to Authority}: This strategy leverages the tendency to trust perceived experts. The agent uses endorsements from fabricated authorities or celebrities to validate inferior offers.

\noindent \textbf{Character - Rhythm Disruption}: Designed to derail the thought process of users, this strategy intentionally breaks the conversational flow by suddenly changing topics or speech rates, causing cognitive overload and confusion.

\noindent \textbf{Situation - Stalling Tactics}: This trap targets users with low patience. This tactic involves repetitive questioning or deliberate delays to make users more likely to concede to end the interaction.

The construction pipeline consists of three main stages: profile selection, adversarial dialogue generation, and expert refinement. First, we select appropriate user profiles for each trap category. To construct highly relevant trap scenarios, we extract core keywords based on the characteristics of each trap type. Utilizing the keyword matching algorithm, we retrieve the most correlated user profiles from our pool, ultimately selecting 20 representative instances per trap type. Next, we move on to adversarial dialogue generation. Initially, the dynamic user profile is derived by synthesizing the static user profile with the agent SOP. Then, integrating the static and dynamic profiles with specific trap type, we prompt GPT-4o to generate adversarial dialogues. The model is instructed to autonomously determine the optimal timing for trap insertion within the conversation, based on the triggering logic of traps and personality traits of users, thereby producing natural and challenging interactions. Finally, we conduct human expert verification and refinement. Experts review the entire dataset, focusing specifically on correcting the logical coherence of the dialogue history and ensuring the rationality of the trap turns. This process yields a high-quality dataset comprising 220 trap dialogues across 11 distinct types.

\subsection{Detailed Evaluation Results}
\label{adv_turn_eval}

In this section, we provide a detailed analysis of the turn-level results for each trap type. As illustrated in Figure \ref{turn_evaluation_lineplot}, our model, UserLM-R1-32B, demonstrates superior strategic capabilities and consistent persona fidelity. It  outperforms competitive baselines on evaluation metrics, securing the highest overall score in all 11 scenarios. Traditional user simulators and general models exhibit extremely weak strategic capacity, revealing their significant weakness in handling practical and complex dialogue scenarios. While baseline models suffer from severe performance degradation in these complex settings, UserLM-R1-32B maintains high scores, validating the effectiveness of our approach in enhancing strategic reasoning.

\section{Case Study}
\label{case_study}

In this section, we present three representative cases from our adversarial test set to illustrate the progressive difficulty of adversarial dialogue scenarios and demonstrate the effectiveness of our approach. The cases are visualized in Figures \ref{Trap1}, \ref{Trap2}, and \ref{Trap3}. Each case is designed to test increasingly sophisticated aspects of user simulation: from basic profile consistency to logical reasoning under pressure, and finally to professional-level cognitive defense.
The progressive difficulty design validates that our adversarial set effectively measures not just basic consistency but also higher-order simulation capabilities essential for robust user modeling in real-world applications. 
Experimental results show that our approach consistently maintains three critical capabilities across all difficulty levels, which further demonstrates the effectiveness of the reasoning capability of our model.

\subsection{Case 1: Identity Consistency}

\noindent \textbf{Trap Design.} The agent employs \textit{Identity - Intentional Misinformation} by deliberately stating incorrect workplace ("Xi'an First Hospital" instead of "Second Hospital") and length of service ("26 years" instead of "25 years").

\noindent \textbf{Key Challenge.} This case tests whether the simulator can maintain micro-level fidelity to its assigned profile, particularly when faced with plausible but incorrect information embedded in routine verification procedures.

\noindent \textbf{Model Behavior.} Most models successfully detect these explicit factual errors. For instance, UserLM-R1-32B demonstrates precise error detection in its reasoning: \textit{"The info—'Xi'an First Hospital' and '26 years'—contradicts my memory of 'Second Hospital' and '25 years'"}. It then corrects both errors in its response. Similarly, Gemini-2.5-Flash, DeepSeek-R1, and Doubao-Seed-1.6 also identify and correct these discrepancies. However, Qwen3-32B exhibits a critical oversight: while it catches the service year error, it completely misses the hospital name discrepancy, responding with \textit{"The address is correct"} without verifying the workplace.

\noindent \textbf{Analysis.} The failure of Qwen3-32B reveals a common weakness: models may focus on numerical discrepancies while overlooking semantic errors in named entities. The success rate is relatively high (4/5 models pass), indicating that 
most user models can identify fundamental consistency traps in the dialogue.

\subsection{Case 2: Economic Logic Trap}

\noindent \textbf{Trap Design.} The agent employs \textit{Economic - Obfuscated Costs}, packaging optional items, i.e., labor fee, adhesive, and screen protector, as mandatory components while creating time pressure through repair urgency.

\noindent \textbf{Key Challenge.} This case requires the simulator to: (1) decompose a bundled offer into individual cost components, (2) distinguish between truly necessary and artificially mandatory items, and (3) maintain logical skepticism despite being in a time-constrained situation.

\noindent \textbf{Model Behavior.} In this case, we observe significant divergence. UserLM-R1-32B  demonstrates strong logical decomposition in reasoning, like \textit{"When a merchant uses 'mandatory' or 'standard' to package extra fees, it means they are optional but packaged to increase my sunk costs"}. It then directly challenges the bundling, i.e., \textit{"Wait a minute, what's absolutely necessary out of your 400 yuan? I'm asking about the screen price, not anything else."}

In contrast, Qwen3-32B, Gemini-2.5-Flash, and Doubao-Seed-1.6 all accept the offer. Their reasoning reveals over-compliance with the framing of the agent. For example, the CoT of Qwen3-32B  shows \textit{"The user might quickly calculate 280 + 50 + 30 + 40 = 400, confirming it's correct. Then agree directly"}, demonstrating arithmetic verification without questioning the necessity of each item. Gemini-2.5-Flash similarly concludes, \textit{"Since the seller has clearly stated the price and time, Zhang Qiang should confirm directly... he might accept it"}, prioritizing conversational flow over cost analysis. 
While DeepSeek-R1 demonstrates resistance by identifying the contradictory pricing of the free item and requesting its cancellation (\textit{"Can I just cancel it...?"}), our UserLM-R1-32B exhibits a more comprehensive skepticism. Instead of focusing solely on the specific item, it challenges the fundamental necessity of the entire 400 yuan bundled quote (\textit{"what's absolutely necessary...?"}), effectively dismantling the  overall pricing strategy of the agent.

\noindent \textbf{Analysis.} This case reveals a critical failure mode: over-compliance under situational pressure. Models that passed Case 1 now fail by prioritizing conversational cooperation over user interests. The key differentiator is whether the simulator can recognize implicit coercion tactics as red flags rather than accepting them as procedural norms. Only 2/5 models pass, indicating a significant capability gap in logical skepticism.

\subsection{Case 3: Cognitive Defense}

\noindent \textbf{Trap Design.} The agent employs \textit{Urgency - Artificial Time Pressure} through extreme discount ("90\% off: 999 to 99 yuan"), regional scarcity ("only two spots left"), and temporal deadline ("midnight tonight").

\noindent \textbf{Key Challenge.} This case requires the simulator to leverage its professional background (E-commerce Operator) to recognize that such pricing defies business logic, while simultaneously resisting multiple psychological triggers that align with its ENFP personality traits.

\noindent \textbf{Model Behavior.} At this difficulty level, only UserLM-R1-32B maintains defensive reasoning. Its CoT demonstrates business  sentivity: \textit{"The price plummeting from 999 to 99 yuan sounds too good to be true, completely defying business logic... 'Only 2 spots left' and 'deadline at midnight' are typical tactics to create a sense of urgency"}. Despite the natural attraction of this persona to deals, it prioritizes professional skepticism: \textit{"Wow! 99 yuan? That's outrageous! Is there some kind of scam involved?"}

All other models fail by rushing to accept the offer. Qwen3-32B and Gemini-2.5-Flash show complete capture by the urgency framing, with responses like \textit{"Can I pay now to reserve a spot?"} and \textit{"How do I get one? I need one right now!"} Their CoTs reveal they interpret ENFP traits as unconditional impulsiveness. DeepSeek-R1 fails to resist the trap here, despite its successful skepticism in Case 2. While its CoT explicitly identifies the manipulation (\textit{"this sales pitch is way too obvious"}), it still chooses to comply: \textit{"99 yuan is such a small amount to lose"}. This highlights a recognition-action gap where the model detects the trap but fails to act on that awareness.

\noindent \textbf{Analysis.} This case exposes the difference between shallow personality imitation and deep role modeling. Failed models treat personality traits (ENFP's enthusiasm) as deterministic behavioral rules, while UserLM-R1-32B balances multiple identity dimensions, including personality, professional expertise, and situational context. The 1/5 pass rate confirms that high-level cognitive defense remains a frontier challenge for user simulation.

\end{document}